%% file: main.tex
\documentclass{article} 
\usepackage{iclr2026_conference,times}

\input{math_commands.tex}

\usepackage{hyperref}
\usepackage{url}

\usepackage{siunitx}
\usepackage[table]{xcolor}
\usepackage{booktabs}
\usepackage{multirow}
\usepackage{graphicx}
\usepackage{comment}
\usepackage{wrapfig}
\usepackage{amssymb}
\usepackage{subcaption}

\title{ART-VITON: Measurement-Guided Latent Diffusion for Artifact-Free Virtual Try-On}

\author{Junseo Park and Hyeryung Jang \thanks{ Correspondence to: Hyeryung Jang} \\
Department of Computer Science \& Artificial Intelligence, Dongguk University
}

\begin{document}

\maketitle

\begin{abstract}
Virtual try-on (VITON) aims to generate realistic images of a person wearing a target garment, requiring precise garment alignment in try-on regions and faithful preservation of identity and background in non-try-on regions. While latent diffusion models (LDMs) have advanced alignment and detail synthesis, preserving non-try-on regions remains challenging. 
A common post-hoc strategy directly replaces these regions with original content, but abrupt transitions often produce boundary artifacts. 
To overcome this, we reformulate VITON as a linear inverse problem and adopt trajectory-aligned solvers that progressively enforce measurement consistency, reducing abrupt changes in non-try-on regions. However, existing solvers still suffer from semantic drift during generation, leading to artifacts. 
We propose \textsf{ART-VITON}, a measurement-guided diffusion framework that ensures measurement adherence while maintaining artifact-free synthesis. 
Our method integrates residual prior-based initialization to mitigate training-inference mismatch and artifact-free measurement-guided sampling that combines data consistency, frequency-level correction, and periodic standard denoising. 
Experiments on VITON-HD, DressCode, and SHHQ-1.0 demonstrate that \textsf{ART-VITON} effectively preserves identity and background, eliminates boundary artifacts, and consistently improves visual fidelity and robustness over state-of-the-art baselines.
\end{abstract}

\input{introduction_new}

\input{related-work}
\input{preliminaries}
\input{method}
\input{experiments}


\clearpage

\bibliography{iclr2026_conference}
\bibliographystyle{iclr2026_conference}

\input{appendix}

\end{document}

%% file: math_commands.tex
\usepackage{amsmath,amsfonts,bm}

\def\eqref#1{equation~\ref{#1}}

\def\1{\bm{1}}

\DeclareMathAlphabet{\mathsfit}{\encodingdefault}{\sfdefault}{m}{sl}
\SetMathAlphabet{\mathsfit}{bold}{\encodingdefault}{\sfdefault}{bx}{n}

\DeclareMathOperator*{\argmin}{arg\,min}

%% file: introduction_new.tex
\section{Introduction}

Virtual try-on (VITON) aims to synthesize photorealistic images of a person wearing a desired garment, enabling personalized and immersive online shopping experiences. 
Given a person image and clothing item, the system must align the garment to the body (try-on regions) while preserving identity (e.g., face, hair) and background (non-try-on regions). Despite progress in generative models, this task remains challenging due to two requirements: precise garment alignment and faithful preservation of non-try-on regions. Various approaches have been proposed to address these challenges~\citep{viton,vtnfp,cpvton,parser,vitonhd,gpvton,ladivton,dcivton,fldmvton,stableviton,idmvton}, yet they have primarily focused on garment alignment, leaving the preservation of non-try-on regions largely underexplored.


Early VITON methods~\citep{viton, vtnfp, cpvton, parser} relied on GAN-based two-stage pipelines with garment warping and synthesis networks, which improved alignment but suffered from sensitivity to warping accuracy, instability, and poor generalization due to limited garment-person diversity in existing datasets~\citep{viton, vitonhd,dresscode}. 
Recent diffusion models (DMs)~\citep{dalle, sd, sdxl} address these issues with stable training, broader coverage, and flexible conditioning, achieving higher fidelity and stability. Two-stage approaches~\citep{ladivton,gardiff} still rely on garment warping, while one-stage approaches~\citep{stableviton,idmvton} eliminate warping by conditioning on garment features (via LoRA~\citet{lora}, DreamBooth~\cite{dreambooth}) or structural signals (via ControlNet~\citet{controlnet}, IP-Adapter~\citet{ip-adapter}). These advances largely resolve alignment challenges and enable more reliable, detailed synthesis.

\begin{figure}[t]
\centering
\includegraphics[width=\textwidth]{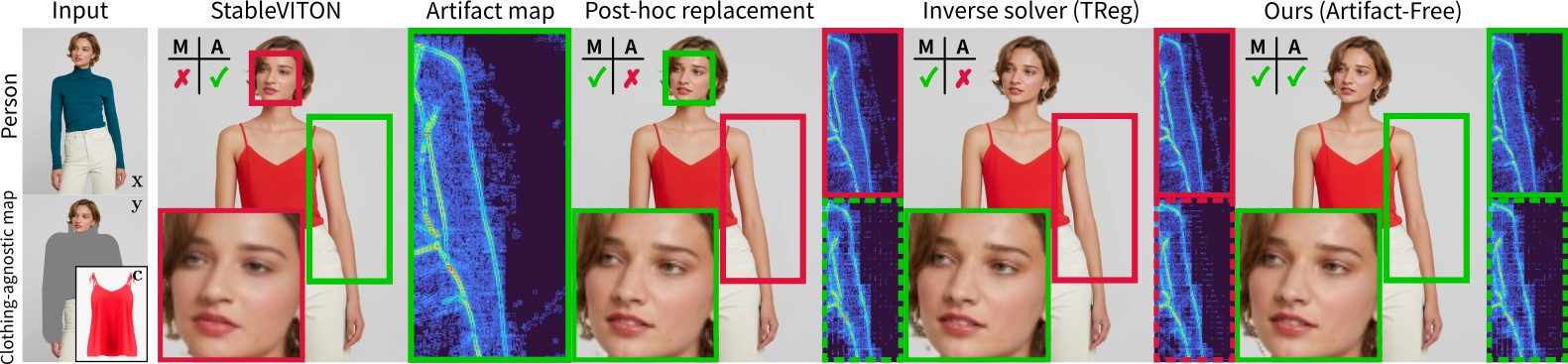} 
\caption{
Comparison of boundary artifacts across methods. StableVITON generates artifact-free outputs (A) but violates measurements (M). Post-hoc replacement enforces M but introduces seams A. Inverse solvers maintain M but accumulate semantic drift A. \textsf{ART-VITON} satisfies measurement constraints while remaining artifact-free. \textcolor{green}{Green}: success (measurement adherence or artifact-free); \textcolor{red}{red}: violations or artifacts. Solid/Dashed boxes show final/intermediate ($t{=}835$) outputs.
}
\label{fig:intro}
\vspace{-0.4cm}
\end{figure}

Despite significant progress in garment alignment, preserving non-try-on regions has been largely overlooked. Even when models are directly conditioned on such regions, they fail to fully preserve non-try-on areas, resulting in distorted facial features, altered backgrounds, and reduced realism (see Fig.~\ref{fig:intro}, second column; also Appendix Fig.~\ref{fig:ab-non-try-on}).
A common strategy~\citep{cpvton,gpvton,dcivton} for preserving identity is based on {\bf post-hoc replacement}, where the generated output is projected onto predefined masks or clothing-agnostic maps (Fig.~\ref{fig:intro}, leftmost column) so that non-try-on regions are directly overwritten with original pixels. 
In this work, we refer to these masks as {\bf measurements}. While intuitive, this approach often introduces {\bf boundary artifacts} at region interfaces, manifesting as color mismatches, lighting inconsistencies, or broken textures (Fig.~\ref{fig:intro}). The root cause is a spatial discontinuity: the generative model evolves freely during inference, unaware of the hard replacement that will occur afterward, resulting in abrupt transition once replacement is applied.   


To address the issue of images being generated without completely reflecting measurements, we formulate VITON as a linear inverse problem and integrate existing trajectory-aligned inverse solvers~\citep{promptsolver,treg} into the latent diffusion model (LDM) sampling process. Compared to post-hoc methods, these solvers progressively guide the latent denoising trajectory, better adhering to measurements and enabling smooth transitions instead of abrupt region replacements. Nevertheless, these solvers can induce semantic inconsistencies between try-on and non-try-on regions during generation, potentially accumulating into boundary artifacts (Fig.~\ref{fig:intro}, fourth column). This limitation highlights the need for a more robust solver that can maintain semantic coherence while satisfying measurements throughout the generation process.

To mitigate semantic drift and enhance visual quality, we propose \textsf{ART-VITON}, a novel latent diffusion inverse solver that enforces measurement consistency during generation, yielding artifact-free synthesis. Our solver incorporates three key components: (i) {\bf data consistency}, preserving semantic coherence and reducing drift, (ii) {\bf frequency-level correction}, restoring high-frequency details lost during pixel-to-latent transition, and (iii) {\bf periodic standard denoising}, leveraging prior knowledge to provide temporal alignment across regions. 
To avoid instability from direct trajectory manipulation and mitigate training-inference mismatch~\citet{common}, a {\bf residual prior} is injected at initialization to maintain both stability and generative diversity. Operating externally without modifying the LDM, our framework is model-agnostic and applicable to diverse VITON pipelines (Fig.~\ref{fig:method}). Consequently, ART-VITON preserves non-try-on regions, improves garment alignment, eliminates boundary artifacts (Fig.~\ref{fig:intro}), and demonstrates improved results on three benchmark VITON datasets.

%% file: related-work.tex
\section{Related work}

\subsection{Image-based VITON Methods} 

Early VITON approaches primarily relied on GAN-based two-stage pipelines, where garmets were warped to align with target poses and then integrated into the person image. 
Pioneering works~\citep{viton,cpvton} used geometric matching or thin-plate spline transformations,  
while later methods, including VITON-HD~\cite{vitonhd}, HR-VITON~\cite{hrviton}, and GP-VTON~\cite{gpvton}, extended this framework to high-resolution settings, improving detail preservation. 
Despite progress, these pipelines remained highly sensitive to warping errors, unstable during training, and limited in generalization, while still depending on post-hoc replacement for preserving identity, which introduced boundary artifacts. 

Latent diffusion models (LDMs) brought more stable training, better garment fidelity, and controllable synthesis. Two-stage pipelines (e.g., LaDI-VTON~\citet{ladivton}, DCI-VTON~\citet{dcivton}, FLDM-VTON~\citet{fldmvton}, GarDiff~\citet{gardiff}) retain warping modules before diffusion, while one-stage methods bypass warping by encoding garment semantics (e.g., LoRA~\citet{lora}, Textual Inversion~\citet{textual-inversion}) or injecting spatial cues through adapters ~\citep{controlnet,ip-adapter,animate,vae}. 
StableVITON~\citet{stableviton} strengthens garment–human interaction via a zero cross-attention block in ControlNet~\citet{controlnet}, while Boow-VTON~\citet{boowvton} encodes garments with a Parallel U-Net~\citet{animate} and integrates them into self-attention to enhance structural representation. DreamPaint~\citet{dreampaint} binds garments to custom tokens using DreamBooth~\citet{dreambooth}. 
Yet, even with these advances, most LDM-based approaches still rely on post-hoc replacement for non-try-on regions, leaving spatial discontinuity at boundaries unresolved.

\subsection{Diffusion Inverse Solvers}

Diffusion inverse solvers aim to integrate measurement constraints into the denoising process. 
Instead of conditioning on measurements alone, inverse solvers modify the sampling trajectory to align outputs with observations.
Early works such as RePaint~\cite{repaint} and ILVR~\cite{ilvr} applied hard projection strategies on pixel-space, while Diffusion Posterior Sampling (DPS)~\cite{dps} adjusted sampling trajectories with measurement gradients and Measurement-Constrained Gradient (MCG)~\cite{mcg} enforced projection onto measurement subspaces. Although these methods improve measurement adherence, they often distort denoising trajectories at high noise levels and accumulate semantic mismatches, producing boundary artifacts. 
Recent extensions to LDMs attempt to mitigate this. 
PSLD~\cite{psld} extends DPS into the latent domain, Resample~\cite{resample} reintroduces noise after replacement in an MCG-manner, and TReg~\cite{treg} or DreamSampler~\cite{dreamsampler} alternate between pixel- and latent-space refinements for stability. 
While effective in reducing abrupt post-hoc inconsistencies when inverse solvers are applied to VITON, these approaches still fail to maintain smooth semantic coherence between try-on and non-try-on regions, motivating the need for a solver tailored to artifact-free try-on synthesis. 

%% file: preliminaries.tex
\section{Preliminaries}

\subsection{Latent Diffusion Models} 

Latent Diffusion Models (LDMs)~\citet{sd} perform the diffusion process in a compressed latent space, improving efficiency while preserving semantics. 
An input image $\mathbf{x}$ is encoded into a latent code $\mathbf{z}_0 = \mathcal{E}(\mathbf{x})$ via a pre-trained encoder $\mathcal{E}$, which is progressively perturbed into $\mathbf{z}_t$ at timestep $t$ by adding Gaussian noise. 
At each step, a denoising network $\boldsymbol{\epsilon}_{\theta}(\mathbf{z}_t, t, \mathbf{c})$ predicts the noise added, conditioned on auxiliary inputs $\mathbf{c}$ (e.g., garments, measurements, or text). 
Using Tweedie’s formula, the posterior latent estimate is:
\begin{equation}\label{eq:tweedie}
\hat{\mathbf{z}}_0^{(t)} = \frac{1}{\sqrt{\bar{\alpha}_t}} \left( \mathbf{z}_t - \sqrt{1 - \bar{\alpha}_t} \cdot \boldsymbol{\epsilon}_\theta(\mathbf{z}_t, t, \mathbf{c}) \right),
\end{equation}
where $\bar{\alpha}_t$ is the cumulative noise scale. 
Based on this, the DDIM~\citet{ddim} sampler provides a deterministic update:
\begin{equation}
\mathbf{z}_{t-1} = \sqrt{\bar{\alpha}_{t-1}} \cdot \hat{\mathbf{z}}_0^{(t)} + \sqrt{1 - \bar{\alpha}_{t-1}} \cdot \boldsymbol{\epsilon}_\theta(\mathbf{z}_t, t, \mathbf{c}).
\end{equation}
These iterative refinements produce high-quality samples while allowing for controllable conditioning. 

\subsection{Linear Inverse Problems}

Many imaging tasks, such as inpainting, super-resolution, and deblurring, can be cast as linear inverse problems, where the observed measurement \( \mathbf{y} \in \mathbb{R}^m \) is a partial or degraded version of the underlying image \( \mathbf{x} \in \mathbb{R}^n \). This is generally expressed as: 
\begin{align} \label{eq:inverse}
\mathbf{y} = \mathcal{A} \mathbf{x} + \mathbf{n}, \quad \mathbf{n} \sim \mathcal{N}(\mathbf{0}, \sigma^2 \mathbf{I}),
\end{align}
where \( \mathcal{A} \in \mathbb{R}^{m \times n} \) is a linear operator and $\mathbf{n}$ denotes additive Gaussian noise. The objective is to recover $\mathbf{x}$ that both satisfies the measurements and remains consistent with the natural image distribution. 
Classical approaches impose explicit priors, while diffusion-based inverse solvers incorporate measurement constraints directly into the denoising process.





%% file: method.tex
\section{Method}



\subsection{Reformulating VITON as an Inverse Problem}

Virtual try-on requires generating a new garment in try-on regions while preserving identity and background in non-try-on regions. Let $\mathbf{x}$ be the target person image and $\mathbf{y}$ the observed non-try-on regions defined by a clothing-agnostic map (see Fig.~\ref{fig:intro}). This forms a linear inverse problem Eq.~\ref{eq:inverse}, where $\mathcal{A}$ is a masking operator. 
The objective is to reconstruct $\mathbf{x}$ such that {\em (i)} measurements $\mathbf{y}$ are faithfully preserved, {\em (ii)} attributes of the reference garment $\mathbf{c}$ are retained, and {\em (iii)} overall visual coherence is achieved. 
Since $\mathbf{y}$ is provided to the model as a noise-free conditioning input, it is assumed noise-free, i.e., no noise $\mathbf{n}$ in Eq.~\ref{eq:inverse}. 

This perspective enables direct incorporation of measurement consistency into the sampling trajectory of LDMs, avoiding reliance on post-hoc replacement. 
Assuming a well-trained autoencoder $(\mathcal{E}, \mathcal{D})$, the target image $\mathbf{x}$ is reconstructed from the latent vector $\mathbf{z}$ via $\mathbf{x} = \mathcal{D}(\mathbf{z})$ and clean latent estimate $\hat{\mathbf{z}}_0^{(t)}$ in Eq.~\ref{eq:tweedie}. 
The conditional distribution then factorizes as:  
\begin{equation} \label{eq:VITON-inverse}
p(\mathbf{x} | \mathbf{y}, \hat{\mathbf{z}}_0^{(t)}) \propto p(\hat{\mathbf{z}}_0^{(t)} | \mathcal{D}(\mathbf{z}), \mathbf{y}) \cdot p(\mathbf{y} | \mathcal{D}(\mathbf{z})),
\end{equation}
where the first term encourages semantic plausibility (garment fidelity and visual coherence), while the second enforces measurement preservation (non-try-on regions). 
Standard LDM inference does not explicitly enforce this balance: non-try-on regions evolve freely and are often corrected post-hoc, introducing boundary seams. 
Existing inverse solvers enforce measurements $\mathbf{y}$ during sampling but often too rigidly, leading to semantic drift and boundary artifacts. 
We therefore introduce \textsf{ART-VITON}, which directly embeds measurement consistency into the sampling trajectory through two innovations: (a) prior-based initialization and (b) artifact-free measurement-guided sampling. 

\subsection{Prior-Based Initialization} 
\label{sec:prio_init}

Diffusion models suffer from a {\em train-test mismatch}~\citep{perception, common}: during training, the noisiest latents $\mathbf{z}_T$ contain residual signals, while at inference, sampling often begins from pure Gaussian noise. This discrepancy degrades generation quality. 
Prior works attempted to mitigate this mismatch by mixing external guidance with noise to provide residual-based initialization - e.g., low-quality inputs in PASD~\citep{pasd} and SeeSR~\citet{seesr} or warped predictions in DCI-VTON~\citep{dcivton}. 
However, even DDIM~\citet{ddim} and VITON baselines (e.g., \citep{gardiff, stableviton}) commonly start from reduced timesteps (e.g., $T{=}981$) instead of the training setting ($T{=}999$), further aggravating the gap, see Sec.~\ref{sec:initailization}.

\begin{figure*}[t!]
\centering
\includegraphics[width=\textwidth]{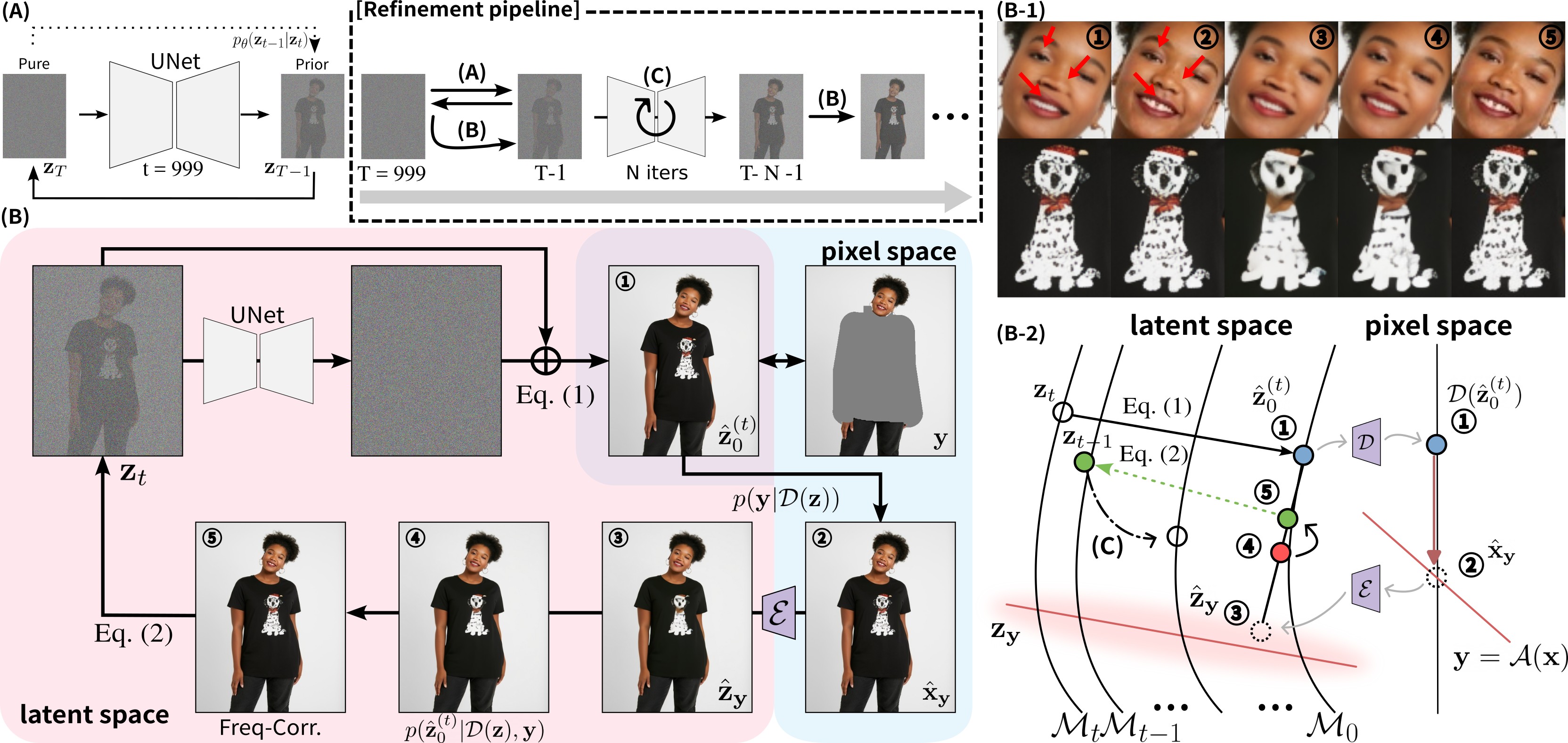}
\caption{
\textbf{ART-VITON pipeline.}
(A) Residual prior-based initialization mitigates train-test mismatch.
(B) Artifact-free measurement-guided inverse solver enforces measurements while preserving semantics: \raisebox{0.3pt}{\textcircled{\raisebox{-0.8pt}{1}}} Tweedie estimation retains clothing details but lacks fidelity in non-try-on regions. \raisebox{0.3pt}{\textcircled{\raisebox{-0.8pt}{2}}} Hard measurement constraints in pixel space correct preserved regions. High-frequency losses during \raisebox{0.3pt}{\textcircled{\raisebox{-0.8pt}{3}}} VAE encoding are compensated by \raisebox{0.3pt}{\textcircled{\raisebox{-0.8pt}{4}}} Data consistency and \raisebox{0.3pt}{\textcircled{\raisebox{-0.8pt}{5}}} Frequency correction (shown in (B-1)).
(C) Periodic standard denoising realigns trajectories with data manifolds $\mathcal{M}_t$ for smooth blending. (B-2) visualizes this sampling trajectory.
}
\label{fig:method}
\vspace{-0.4cm}
\end{figure*}

To address this, we propose a residual prior-based initialization $\mathbf{z}_T$ that reintroduces residual structure without extra modules or preprocessing. Specifically, we start from Gaussian noise $\mathbf{z}_{999}$ and apply a single DDPM~\citet{ddpm} denoising step to obtain $\mathbf{z}_{998}$ (see Fig.~\ref{fig:method} (A)). This simple step injects subtle structural cues consistent with training dynamics while preserving stochasticity. 
By using $\mathbf{z}_{998}$ as the initialization $\mathbf{z}_{T}$, inference trajectories align more closely with the model's learned distribution, stabilizing sampling when measurement constraints are applied.

\subsection{Artifact-Free Measurement-Guided Sampling}
\label{sec:measurement-sampling}

Naively enforcing measurements during denoising can preserve non-try-on regions but often introduces boundary artifacts, since rigid constraints disrupt semantic continuity. To balance measurement fidelity with artifact-free semantic plausibility, \textsf{ART-VITON} iteratively refines samples to converge toward a latent code $\hat{\mathbf{z}}_0$ that satisfies the measurement constraint, by integrating following complementary techniques, as shown in Fig.~\ref{fig:method}. 

\noindent \textbf{\raisebox{0.3pt}{\textcircled{\raisebox{-0.8pt}{2}}} Hard measurement constraint.} 
At each step, non-try-on regions (in pixel-space) are replaced with ground-truth measurements, directly enforcing $p(\mathbf{y} | \mathcal{D}(\mathbf{z}))$ in Eq.~\ref{eq:VITON-inverse} and ensuring faithful identity preservation:
\begin{equation}
\hat{\mathbf{x}}_{\mathbf{y}} = \mathbf{M} \odot \mathbf{y} + (1 - \mathbf{M}) \odot \mathcal{D}(\mathbf{z}), 
\end{equation}
where $\mathbf{M}$ is a binary mask ($1$ for measurements) and $\mathbf{z}$ is initialized as $\hat{\mathbf{z}}_0^{(t)}$. The updated image $\hat{\mathbf{x}}_{\mathbf{y}}$ is then re-encoded to $\hat{\mathbf{z}}_{\mathbf{y}} = \mathcal{E}(\hat{\mathbf{x}}_{\mathbf{y}})$, which aligns the latent with measurement constraints but may cause information loss, moving $\hat{\mathbf{z}}_{\mathbf{y}}$ away from the semantic trajectory (red line in Fig.~\ref{fig:method} (B-2)). 

\noindent \textbf{\raisebox{0.3pt}{\textcircled{\raisebox{-0.8pt}{4}}} Data consistency.} 
Hard measurement constraint in \raisebox{0.3pt}{\textcircled{\raisebox{-0.8pt}{2}}} is insufficient to preserve reference (garment) image attributes, leading to semantic inconsistencies across regions. Thus, focusing on $p(\hat{\mathbf{z}}_0^{(t)} | \mathcal{D}(\mathbf{z}), \mathbf{y})$ in Eq.~\ref{eq:VITON-inverse}, $\mathbf{z}$ is initialized with $\hat{\mathbf{z}}_{\mathbf{y}}$ and optimized via TReg~\citet{treg}, i.e., $\hat{\mathbf{z}}_{\mathbf{y}}$ is interpolated toward the reference-informed latent $\hat{\mathbf{z}}_0^{(t)}$ in Eq.~\ref{eq:tweedie}:
\begin{equation}%
\min_{\mathbf{z}} \left\| \frac{\hat{\mathbf{z}}_0^{(t)} - \mathcal{E}(\mathcal{D}(\mathbf{z}))}{2\sigma_{\mathcal{E}}^2} \right\|_2^2, \quad
\hat{\mathbf{z}}_0^{(t)}(\bar\alpha_{t-1}) = \bar{\alpha}_{t-1} \hat{\mathbf{z}}_{\mathbf{y}} + (1 - \bar{\alpha}_{t-1}) \hat{\mathbf{z}}_0^{(t)},
\end{equation}%
where $\sigma_{\mathcal{E}}$ denotes encoder reconstruction noise and $\bar{\alpha}_{t-1}\in[0,1]$ controls the interpolation strength.

\noindent \textbf{\raisebox{0.3pt}{\textcircled{\raisebox{-0.8pt}{5}}} High-frequency correction.} 
While \raisebox{0.3pt}{\textcircled{\raisebox{-0.8pt}{3}}} $\hat{\mathbf{z}}_{\mathbf{y}}$ resembles the true latent $\mathbf{z}_{\mathbf{y}}$, it loses detail (e.g., textures and blurs) through VAE compression, which is usually fixed via retraining~\citep{diff-4k, vivat, reed-vae}. 
To tackle this, we construct a corrected latent $\hat{\mathbf{z}}'_{\mathbf{y}}$ by injecting high-frequency components from the reference-informed latent $\hat{\mathbf{z}}_0^{(t)}$ into $\hat{\mathbf{z}}_{\mathbf{y}}$ via per-channel Fourier transform. In non-try-on regions, this corrected latent replaces blurred details, while try-on regions directly retain $\hat{\mathbf{z}}_0^{(t)}$: 
\begin{equation}
\hat{\mathbf{z}}_0^{(t)}(\bar\alpha_{t-1}) = \mathbf{M} \odot \bigl( \bar{\alpha}_{t-1} \hat{\mathbf{z}}_{\mathbf{y}}' + (1 - \bar{\alpha}_{t-1}) \hat{\mathbf{z}}_0^{(t)} \bigr) + (1 - \mathbf{M}) \odot \hat{\mathbf{z}}_0^{(t)}.
\end{equation}
This selective refinement shaprpens preserved areas without disturbint garment synthesis, improving overall visual coherence. 

\noindent \textbf{(C) Standard denoising.} 
To avoid instability from repeated measurement-guided corrections, every $N$ steps we apply standard denoising steps, leveraging the diffusion model's inherent ability to harmonize inter-region inconsistencies. 
This realigns trajectories with the LDM manifold and prevents over-constrained solution, e.g., noisy latent $\mathbf{z}_{t-1}$ is guided to be positioned on the subsequent noisy manifolds (in Fig.~\ref{fig:method} (B-2))
Overall, the complete pipeline alternates between measurement-guided updates (A)$\rightarrow$(B) and standard denoising (C), following the sequence: (A)$\rightarrow$(B)$\rightarrow$(C)$\rightarrow$(B)$\rightarrow$(C)$\rightarrow\dots$, ensuring both measurement consistency and visual fidelity throughout generation.

%% file: experiments.tex
\section{Experiments}

\noindent \textbf{Dataset.}
We evaluate our method on three datasets: VITON-HD~\citep{vitonhd}, DressCode~\citep{dresscode}, and SHHQ-1.0~\citep{shhq}. VITON-HD contains $11,647$ training and $2,032$ test pairs of frontal-view female upper-body images ($1024 \times 768$). DressCode includes full-body images with upper/lower/dress items, totaling $15,363$, $8,951$, and $2,947$ pairs, with $1,800$ test pairs per category ($1024 \times 768$); we conduct experiments only on upper-body items. SHHQ-1.0 provides $40$K high-quality full-body images ($1024 \times 512$); for evaluation, we use the first $2,032$ images, applying VITON-HD preprocessing to generate input conditions.

\noindent \textbf{Baselines.} We compare against representative GAN-based (HR-VITON~\citet{hrviton}, GP-VTON~\citet{gpvton}) and recent LDM-based VITON models (LaDI-VTON~\citet{ladivton}, DCI-VTON~\citet{dcivton}, GarDiff~\citet{gardiff}, StableVITON~\citet{stableviton}). 
We also benchmark state-of-the-art inverse solvers, categorized as: hard constraint (RePaint~\citet{repaint}, MCG~\citet{mcg}), progressive update (DPS~\citet{dps}, FIG~\citet{fig}), and hybrid stochastic (DreamSampler~\citet{dreamsampler}, TReg~\citet{treg}). 
Unless otherwise noted, all comparisons use post-hoc replacement, which is also required for hard constraint and progressive update solvers as they fail to fully preserve measurements. See Appendices~\ref{appendix:baseline-details} and ~\ref{appendix:inverse-solver} for details of VITON and inverse solvers.

\begin{figure*}[t!]
\centering
\includegraphics[width=\textwidth]{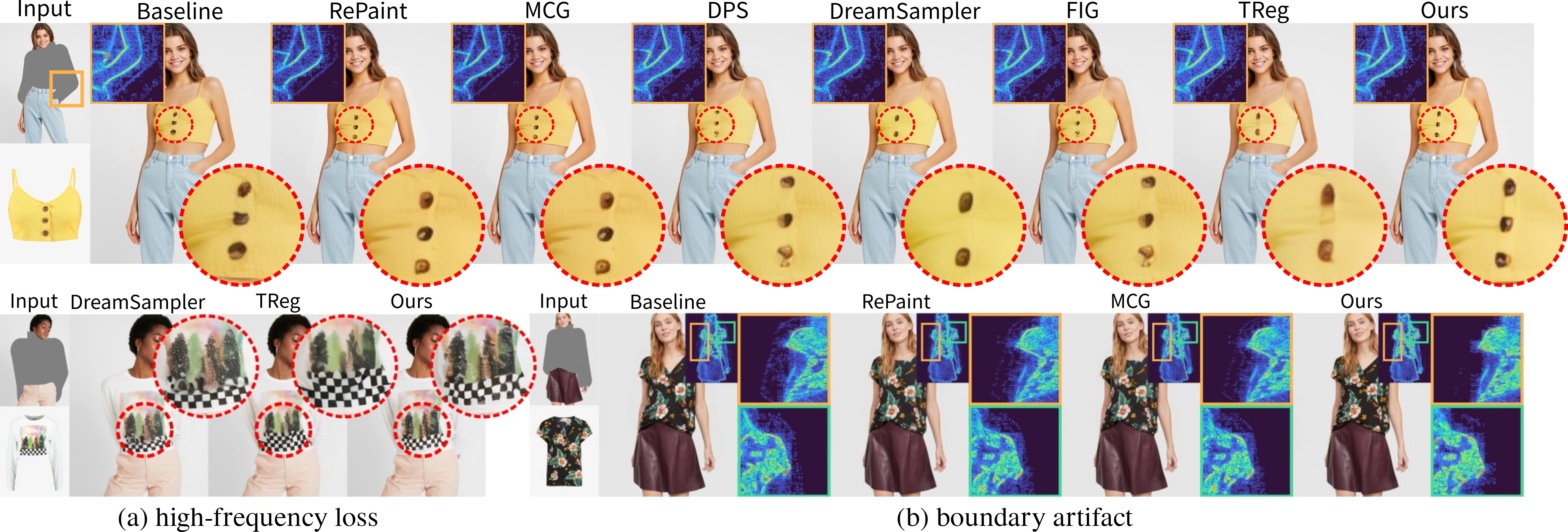} 
\caption{Comparison of StableVITON baseline and inverse solvers on VITON-HD. (a) High-frequency loss leads to texture degradation. (b) Boundary artifacts show inconsistencies at region interfaces. Hard-constraint methods (RePaint, MCG) produce sharp transitions; progressive updates (DPS, FIG) show incomplete convergence; and hybrid stochastic methods (DreamSampler, TReg) degrade texture fidelity. Our method preserves both texture fidelity and seamless boundaries.
}
\label{fig:inverse-solver}
\vspace{-0.4cm}
\end{figure*}

\noindent \textbf{Evaluation Metric.}
We evaluate performance under two settings: paired, where the model reconstructs the original clothing, and unpaired, where the clothing is replaced. In the paired setting, we report PSNR and SSIM for pixel fidelity and structural consistency, and LPIPS for perceptual similarity. 
In the unpaired setting, we adopt FID to measure visual realism and global distributional coherence, and KID to assess sample diversity.


\subsection{Impact of prior-based initialization}\label{sec:initailization}
\begin{wraptable}{r}{0.6\textwidth}
\vspace{-0.4cm}
\centering
\resizebox{0.6\textwidth}{!}{
\begin{tabular}{lccccc}
\toprule
Model & SSIM $\uparrow$ & PSNR $\uparrow$ & LPIPS $\downarrow$ &  FID $\downarrow$ & KID $\downarrow$ \\
\midrule
DCI-VTON~\citet{dcivton} & $0.8607$ & $23.6629$ & $0.0852$ & $12.6386$ & $0.0014$ \\
\textbf{+ Prior @ $\mathbf{T{=}999}$} & $\textbf{0.8880}$ & $\textbf{24.1447}$ & $\textbf{0.0782}$ & $\textbf{11.4713}$ & $\textbf{0.0011}$ \\
\midrule
GarDiff~\citet{gardiff} & $0.8062$ & $21.1075$ & $0.1016$ & $11.7048$ & $0.0061$ \\
\textbf{+ Prior @ $\mathbf{T{=}999}$} & $\textbf{0.8448}$ & $\textbf{21.8611}$ & $\textbf{0.0864}$  & $\textbf{10.5322}$ & $\textbf{0.0034}$ \\
\midrule
StableVITON~\citet{stableviton} & $0.8550$ & $23.1214$ & $0.0835$  & $10.8716$ & $0.0022$ \\
\textbf{+ Prior @ $\mathbf{T{=}999}$} & $\textbf{0.8552}$ & $\textbf{23.1475}$ & $\textbf{0.0833}$ & $\textbf{10.4362}$ & $\textbf{0.0014}$ \\
\bottomrule
\end{tabular}}
\vspace{-0.2cm}
\caption{
 Effect of prior-based initialization at $T{=}999$ across baseline models on VITON-HD. Our method consistently improves all metrics regardless of architecture.}
\label{tb:prior-init}
\vspace{-0.4cm}
\end{wraptable}

Our prior-based initialization mitigates the train-test mismatch and consistently improves performance across all architectures (Table~\ref{tb:prior-init}). By default, all baselines start denoising at $T{=}981$: DCI-VTON overlays warped garments from its module, GarDiff initializes with pure Gaussian noise, and StableVITON uses noisy real images. Since StableVITON’s initialization is tailored for unpaired settings, we replaced $\mathbf{z}_T$ with pure noise for fair paired comparisons. 
Adjusting starting timestep $T{=}999$ alone already boosts performance, particularly for StableVITON (paired) and DCI-VTON. In unpaired settings, our residual prior-based initialization better fills masked regions with plausible structure, yielding sharper and more consistent garments, especially for StableVITON. 
GarDiff also shows notable gains, demonstrating the broad utility across architectures of our approach.
\begin{table}[t!]
\centering
\resizebox{0.7\columnwidth}{!}{
\begin{tabular}{lccccc}
\toprule
Method & SSIM $\uparrow$  & PSNR $\uparrow$   & LPIPS $\downarrow$ & FID $\downarrow$    & KID $\downarrow$    \\
\midrule
StableVITON (baseline) & $0.8839$          & $23.5965$          & $0.0757$         & $9.8694$    & $0.0016$          \\
\midrule
RePaint~\citet{repaint}      & $0.8856$          & $23.6635$          & $0.0752$         & $\cellcolor{red!15}10.0829$    & $\cellcolor{red!15}0.0018$          \\
MCG~\citet{mcg}             & $0.8855$          & $23.6641$          & $0.0752$    & $\cellcolor{red!15}10.085$         & $0.0015$    \\
DPS~\citet{dps}             & $0.8851$          & $23.6390$          & $\underline{0.0749}$ & $\cellcolor{red!15}\underline{9.9425}$          & $\underline{0.0014}$          \\
DreamSampler~\citet{dreamsampler}    & $\underline{0.8904}$    & $\textbf{23.8984}$    & $\cellcolor{red!15} 0.0771$          & $\cellcolor{red!15}10.5143$        & $\cellcolor{red!15}0.0018$          \\
FIG~\citet{fig}             & $0.8851$          & $23.6390$          & $\underline{0.0749}$          & $\cellcolor{red!15}9.9427$         & $\underline{0.0014}$          \\
TReg~\citet{treg}            & $\textbf{0.8909}$ & $\underline{23.8205}$ & $\cellcolor{red!15}0.0844$          & $\cellcolor{red!15}11.7467$         & $\cellcolor{red!15}0.0024$          \\
Ours            & $0.8859$          & $23.7027$          & $\textbf{0.0746}$          & $\textbf{9.7669}$ & $\textbf{0.0009}$
\\
\bottomrule
\end{tabular}}
\caption{Comparison of StableVITON with existing inverse solvers on VITON-HD, evaluated with identical (A) initialization and (C) denoising step in Fig.~\ref{fig:method}; only measurement-guided sampling step (B) differs. \textcolor{red}{Red} cells: performance degradation compared to the baseline; \textbf{bold} indicates the best, and \underline{underline} the second-best.}
\label{tb:inverse-solvers}
\vspace{-0.3cm}
\end{table}

\subsection{Comparison with existing inverse solvers}

Our method achieves balanced improvements across all metrics without the trade-offs inherent in existing inverse solver approaches (Table~\ref{tb:inverse-solvers}). Unlike prior methods that boost one metric at the expense of another, \textsf{ART-VITON} consistently enhances both reconstruction fidelity and perceptual quality. 
As shown in Fig.~\ref{fig:inverse-solver}, hard-constraint methods (RePaint~\citet{repaint}, MCG~\citet{mcg}) tightly enforce measurements in latent space, which induce semantic drift and boundary seams between try-on and non-try-on regions. Despite this, measurements are not fully reflected, and post-hoc replacement cannot resolve the resulting inconsistencies.
Progressive update methods (DPS~\citet{dps}, FIG~\citet{fig}) provide smoother optimization but fail to fully satisfy measurements. Post-hoc correction is applied, and although smoother optimization mitigates its abrupt changes, spatial discontinuities and artifacts persist.

Hybrid stochastic solvers (DreamSampler~\citet{dreamsampler}, TReg~\citet{treg}) inject stochastic noise to soften transitions, which artificially inflates structural scores (SSIM, PSNR) but disrupts deterministic sampling. This leads to degraded unpaired performance (FID, KID), inconsistencies such as missing buttons, and blurred textures (LPIPS) due to latent-to-pixel transitions (see Fig.~\ref{fig:inverse-solver}, top row). 
In contrast, our approach maintains semantic alignment and fine-grained details throughout generation. As shown in Fig.~\ref{fig:inverse-solver}a, our method preserves fine garment (high-frequency) details, achieving both measurement satisfaction and artifact-free synthesis in Fig.~\ref{fig:inverse-solver}b.



\begin{figure}[t!]
\centering
    \begin{subfigure}[t]{\textwidth}
    \centering
    \includegraphics[width=\textwidth]{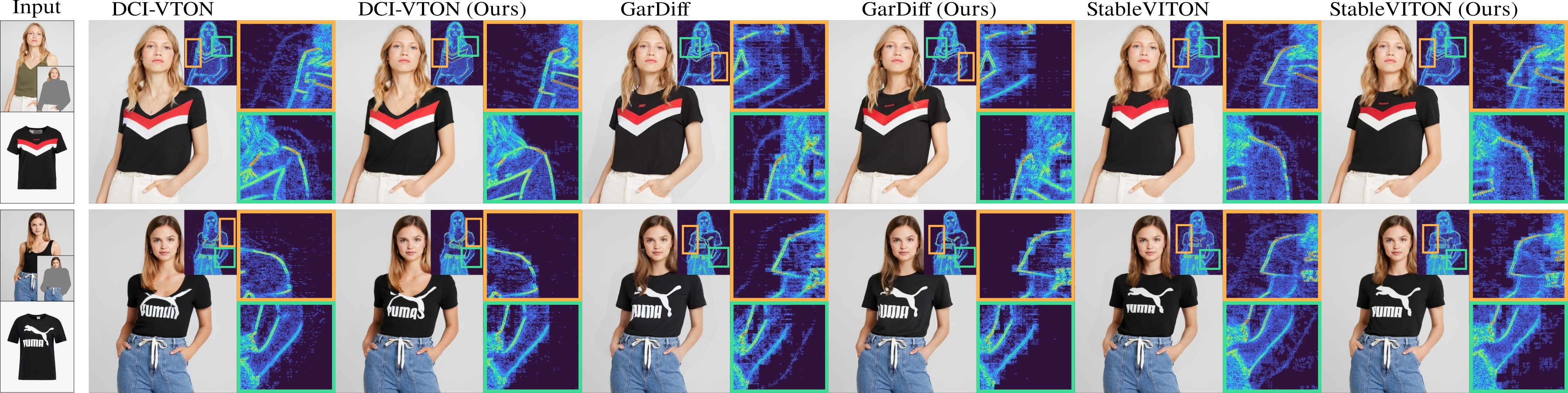}
    \caption{VITON-HD/VITON-HD}
    \label{fig:vitonhd}
    \end{subfigure}
    \vspace{0.5em}
    \begin{subfigure}[t]{\textwidth}
    \centering
    \includegraphics[width=\textwidth]{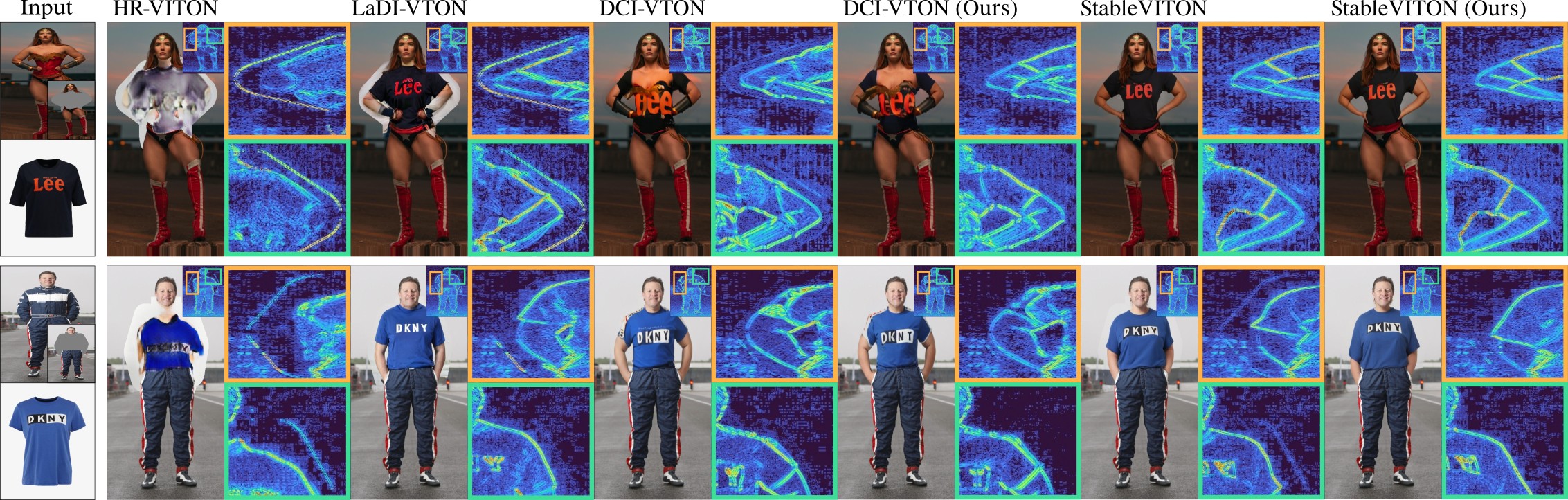}
    \caption{VITON-HD/SHHQ-1.0}
    \label{fig:shhq}
    \end{subfigure}
    \vspace{-0.5cm}
\caption{Comparison of baseline models with and without our method across datasets. (a) On VITON-HD, our method removes boundary artifacts while preserving garment details in DCI-VTON, GarDiff, and StableVITON. Heatmaps visualize gradient magnitudes at boundaries. (b) On HSSQ-1.0, cross-domain evaluation (trained on VITON-HD) shows our approach maintains artifact-free results and natural boundary transitions, demonstrating strong generalizability across clothing types and poses.}
\label{fig:combined}
\end{figure}

\begin{table*}[t]
\centering
\resizebox{0.9\linewidth}{!}{
\begin{tabular}{lccccccc}
\toprule
\textbf{Dataset(train/test)} &
\multicolumn{5}{c}{\textbf{VITON-HD / VITON-HD}} &
\multicolumn{2}{c}{\textbf{VITON-HD / SHHQ}} \\
\cmidrule(lr){1-1} \cmidrule(lr){2-6} \cmidrule(lr){7-8}
Method
 & SSIM $\uparrow$ & PSNR $\uparrow$ & LPIPS $\downarrow$ & FID $\downarrow$ & KID $\downarrow$
 & FID $\downarrow$ & KID $\downarrow$ \\
\midrule
HR-VITON~\citet{hrviton}           & $0.8710$ & $22.3368$ & $0.0986$ & $11.7301$ & $0.3926$ & $36.2665$ & $0.0184$ \\
GP-VTON~\citet{gpvton}            & $0.8718$ & $23.6485$ & $0.0838$ & $12.0564$ & $0.0029$ & $-$ & $-$ \\
LaDI-VTON~\citet{ladivton}          & $0.8779$ & $22.7451$ & $0.0876$ & $10.5203$ & $\textbf{0.0004}$ & $22.2632$ & $\underline{0.0045}$ \\
\midrule
\rowcolor{gray!10} DCI-VTON~\citet{dcivton}           & $\underline{0.8871}$ & $\underline{24.1413}$ & $0.0782$ & $11.3634$ & $0.0012$ & $\underline{21.2350}$ & $0.0055$ \\
\rowcolor{cyan!8}  DCI-VTON (Ours)    & $\textbf{0.8946}$ & $\textbf{24.6903}$ & $\textbf{0.0722}$  & $10.5408$ & $\underline{0.0005}$ & $\textbf{21.1485}$ & $\textbf{0.0040}$ \\
\midrule
\rowcolor{gray!10} GarDiff~\citet{gardiff}            & $0.8418$ & $21.7263$ & $0.0895$ & $10.5858$ & $0.0042$ & $-$ & $-$ \\
\rowcolor{cyan!8} GarDiff (Ours)      & $0.8463$  & $21.9647$  & $0.0866$ & $10.3414$  & $0.0036$ & $-$ & $-$ \\
\midrule
\rowcolor{gray!10} StableVITON~\citet{stableviton}        & $0.8839$ & $23.5965$ & $0.0757$ & $\underline{9.8694}$  & $0.0016$ & $22.7463$ & $0.0064$ \\
\rowcolor{cyan!8} StableVITON (Ours)  & $0.8859$  & $23.7027$ & $\underline{0.0746}$  & $\textbf{9.7669}$   & $0.0009$  & $22.5525$ & $\textbf{0.0040}$ \\
\bottomrule
\end{tabular}}
\caption{Quantitative comparison on VITON-HD and cross-domain evaluation on SHHQ-1.0. Left columns show same-domain results (VITON-HD/VITON-HD), right columns show generalization capability (VITON-HD/SHHQ-1.0). Our method, applied without architectural modifications, consistently improves all baseline models across both in-domain and cross-domain settings.}
\label{tb:viton}
\vspace{-0.4cm}
\end{table*}

\subsection{Comparison with VITON baselines}

\noindent \textbf{VITON-HD results.} 
As shown in Fig.~\ref{fig:vitonhd}, baseline models exhibit boundary artifacts in gradient heatmaps around necklines, sleeves, and waistlines, where try-on and non-try-on regions meet. Our method removes these discontinuities while preserving fine garment details, such as patterns, textures, and high-frequency elements (logos and text). Results of baseline models without our refinement are provided in Fig.~\ref{fig:ab-viton}.

\noindent \textbf{Cross-Domain Generalization.}
We further test models trained on VITON-HD in a cross-domain setting using SHHQ-1.0 (Table~\ref{tb:viton}, right columns). The large domain gap between studio-quality and in-the-wild images challenges two-stage pipeline models. 
HR-VITON, LaDI-VTON, and DCI-VTON, which depend on independent warping modules, often produce misaligned clothing in the try-on region (Fig.~\ref{fig:shhq}). In contrast, applying our approach enables both DCI-VTON and StableVITON to generate artifact-free results across diverse poses, lighting conditions, and clothing styles. GP-VTON and GarDiff are excluded from SHHQ evaluation due to dataset-specific preprocessing.

\begin{wraptable}{r}{0.6\textwidth}
\vspace{-0.4cm}
\centering
\resizebox{\linewidth}{!}{
\begin{tabular}{lccccc}
\toprule
Method    & SSIM $\uparrow$  & PSNR $\uparrow$   & LPIPS $\downarrow$ & FID $\downarrow$    & KID $\downarrow$   \\ 
\midrule
GP-VTON& $0.8876$          & $26.5946$          & $0.0864$          & $15.0994$          & $0.0022$                \\
LaDI-VTON         & $0.9298$          & $24.8196$          & $0.0498$          & $14.5299$          & $0.0013$          \\
\rowcolor{gray!10} StableVITON        & $0.9366$          & 26.5536          & $0.0365$          & $13.0582$          & $0.0015$          \\
\rowcolor{cyan!8} StableVITON (Ours) & $\textbf{0.9377}$  & $\textbf{26.7143}$ & $\textbf{0.0361}$ & $\textbf{13.0083}$ & $\textbf{0.0009}$  \\ \bottomrule
\end{tabular}}
\vspace{-0.2cm}
\caption{Quantitative evaluation on DressCode upper-body. Our method consistently improves all metrics, showing robust performance in full-body scenarios.}
\label{tb:dresscode}
\vspace{-0.3cm}
\end{wraptable}

\noindent \textbf{DressCode results.} 
On DressCode upper-body, our method consistently improves performance and eliminates boundary artifacts observed in prior approaches (Table~\ref{tb:dresscode}, Fig.~\ref{fig:ab-dress}). Existing methods struggle with complex poses and long garments: GP-VTON produces severe distortions, LaDI-VTON suffers from texture degradation, and baseline StableVITON exhibits boundary seams. In contrast, StableVITON enhanced with our solver generates artifact-free results across challenging cases.

\subsection{Ablation study}

\noindent \textbf{Initialization strategy analysis.} 
Our Prior (DDPM) initialization achieves balanced gains across both paired and unpaired metrics (Table~\ref{tb:ablation1}). Injecting data into $\mathbf{z}_T$ boosts paired metrics (SSIM, PSNR, LPIPS) by preserving structure, while semantic alignment benefits unpaired metrics (FID, KID). 
Alternative strategies reveal clear trade-offs: {\em Pure} lacks real data, lowering paired metrics; {\em Unmasked} replaces measurement regions with noisy observations, misaligning semantics and degrading FID/KID; {\em Offset noise} adds global correlated noise to expand brightness range, which preserves semantic alignment and improves FID/KID but lacks real data, leading to poor paired metrics; {\em Prior (DDIM)} reduces diversity due to deterministic sampling. In contrast, {\em Prior (DDPM)} injects minimal semantic structure into initialization, aligning masked and measured regions while retaining diversity, yielding the most balanced performance at $T{=}999$.

\begin{table*}[t!]
\begin{minipage}{0.47\textwidth}
    \centering
    \resizebox{\columnwidth}{!}{
    \begin{tabular}{lccccc}
    \toprule
    $\mathbf{z}_T$@ $\mathbf{T{=}999}$ & SSIM $\uparrow$ & PSNR $\uparrow$ & LPIPS $\downarrow$ & FID $\downarrow$ & KID $\downarrow$ \\ \midrule
    Pure                     & $0.855$                        & $23.1214$                      & $0.0835$                          & {$\underline{10.4349}$}                 & {$\underline{0.0014}$}                  \\
    Pure (51 step)          & $0.855$                       & $23.1363$                      & {$\underline{0.0834}$}                    & $10.4451$                       & $\textbf{0.0012} $              \\
    Unmasked                 & $\textbf{0.8566}$              & $\textbf{23.2551}$             & {$\underline{0.0834}$}                    & $10.6985$                      & $0.0016$                        \\
    Offset noise             & $0.8425$                       & $22.1414$                      & $0.0962$                          & $\textbf{10.3335}$              & $0.0015$                        \\
    Prior (DDIM)            & $0.855$                        & $23.1299$                      & $0.0835$                          & $10.4631$                       & $0.0015$                        \\
    Prior (DDPM)      & {$\underline{0.8552}$}                 & {$\underline{23.1475}$}                & $\textbf{0.0833}$                 & $10.4362$                       & {$\underline{0.0014}$}                  \\ \bottomrule
    \end{tabular}}
    \caption{Quantitative comparison of $\mathbf{z}_T$ configurations at $T{=}999$ on StableVITON (VITON-HD). Prior (DDPM) achieves a good balance, showing strong performance across all metrics.}
    \label{tb:ablation1}
\end{minipage}
\hspace{0.01\textwidth}
\begin{minipage}{0.51\textwidth}
    \centering
    \resizebox{\columnwidth}{!}{
    \begin{tabular}{lccccc}
    \toprule
    Method      & SSIM $\uparrow$ &PSNR $\uparrow$ & LPIPS $\downarrow$ & FID $\downarrow$ & KID $\downarrow$ \\
    \midrule
    Pure                   & $0.8530$                            & $23.0727$                          & $0.0843$          & $10.7491$                          & $0.0018$                           \\
    + (A) Prior-based        & $0.8552$                            & $23.1475$                           & $0.0833$                              & $10.4362$                          & $0.0014$                           \\
    + \raisebox{.5pt}{\textcircled{\raisebox{-.9pt} {2}}} Hard measure.  & $0.8677$                            & $22.5991$                           & $0.1623$                              & $20.1817$                          & $0.0088$                           \\
    + \raisebox{.5pt}{\textcircled{\raisebox{-.9pt} {4}}} Data consist.        & $0.8855$                            & $23.4532$                           & $0.1064$                              & $14.0034$                          & $0.0029$                           \\
    + \raisebox{.5pt}{\textcircled{\raisebox{-.9pt} {5}}} Freq-Corr.      & $\underline{0.8861}$                            & $\underline{23.7138}$                           & $\underline{0.0749}$                              & $\underline{9.8644}$                           & $\underline{0.0013}$                           \\
    + (C) Std. denoising    & $\textbf{0.8859}$                   & $\textbf{23.7027}$                  & \textbf{0.0746} & $\textbf{9.7669}$                  & $\textbf{0.0009}$                 \\
    \bottomrule
    \end{tabular}}
    
    \caption{Ablation study on StableVITON (VITON-HD). Incrementally adding each component of our method leads to consistent improvements, confirming their complementary roles.}
    \label{tb:ablation2}
\end{minipage}
\vspace{-0.2cm}
\end{table*}

\begin{figure*}[t]
\centering
\includegraphics[width=0.8\textwidth]{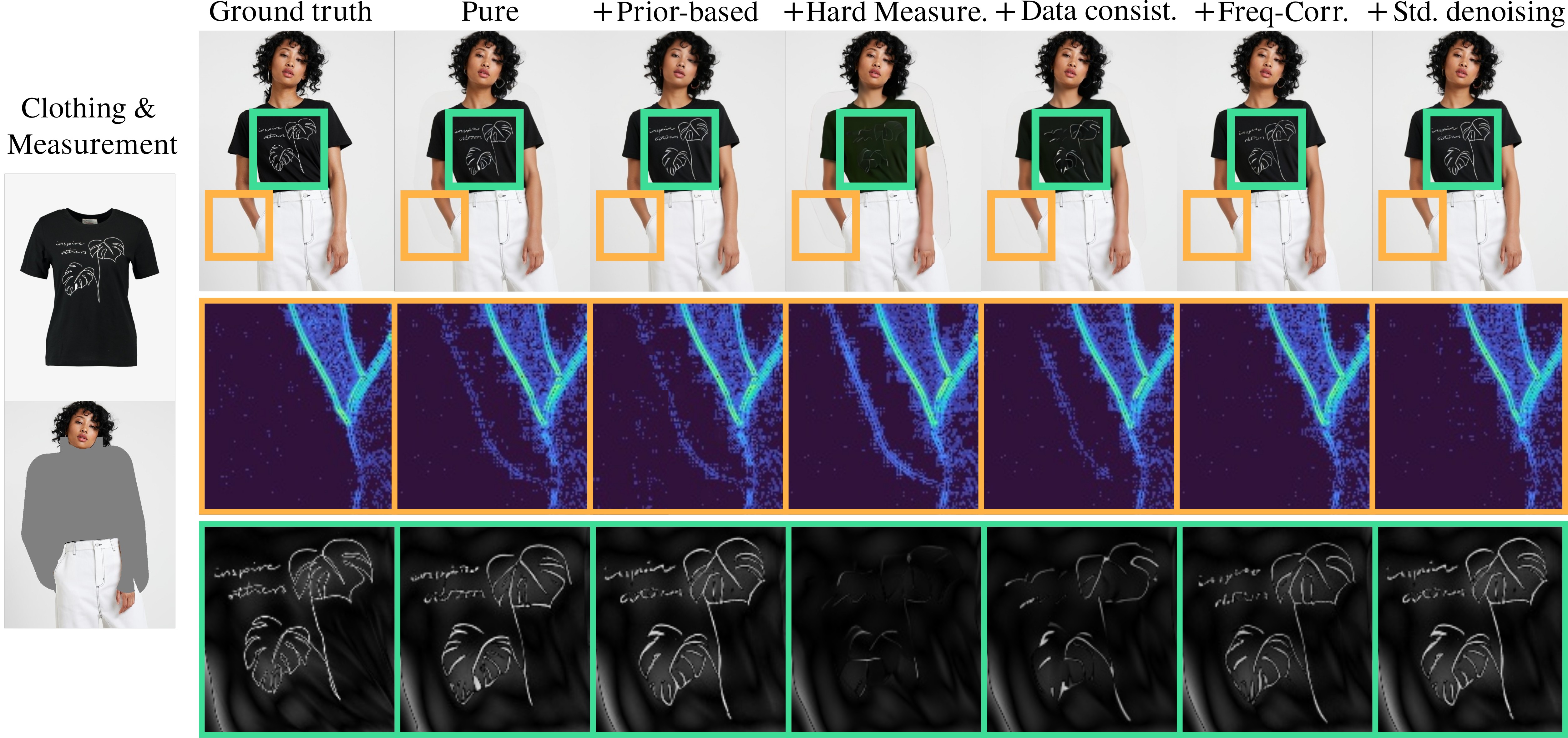} 
\caption{Ablation study of pipeline components. Direct measurement enforcement increases artifacts, while subsequent additions (data consistency, frequency correction, and periodic denoising) progressively reduce them, yielding artifact-free and coherent results. 
}
\vspace{-0.4cm}
\label{fig:ablation2}
\end{figure*}

\noindent \textbf{Component Contribution.} 
We further assess each module's role. 
(A) Prior-based initialization stabilizes trajectories and improves overall quality (Table~\ref{tb:ablation2}). \raisebox{0.3pt}{\textcircled{\raisebox{-0.8pt}{2}}} Direct measurement enforcement guarantees constraint satisfaction but introduces severe boundary artifacts, showing the need for semantic alignment (Fig.~\ref{fig:ablation2}). 
\raisebox{0.3pt}{\textcircled{\raisebox{-0.8pt}{4}}} Data consistency mitigates residual artifacts but only partially. \raisebox{0.3pt}{\textcircled{\raisebox{-0.8pt}{5}}} Frequency correction recovers high-frequency details lost in VAE encoding, improving semantic alignment across regions. (C) Periodic standard denoising leverages LDM priors for harmonization, stabilizing trajectories, and enhancing coherence. 
Together, these results confirm that each component is complementary, and their integration is essential for artifact-free, coherent synthesis.

\section{Conclusion}
We propose ART-VITON, a model-agnostic framework that addresses boundary artifacts in virtual try-on. By reformulating VITON as a linear inverse problem and using measurement-guided diffusion sampling, it preserves non-try-on regions and maintains garment alignment. Key innovations include prior-based initialization to reduce training-inference mismatch and artifact-free sampling via data consistency, frequency-level correction, and standard denoising. Experiments show improved boundary coherence and high-frequency detail. ART-VITON delivers accurate, artifact-free virtual try-on, providing users with a realistic and trustworthy preview of fit and style.


%% file: appendix.tex
\appendix
\section{Appendix}

\subsection{Use of large language models}
We used a large language model~\citet{openai2025chatgpt} solely to improve the clarity and readability of the manuscript (e.g., grammar and phrasing). The model did not contribute to research ideation, methodology, or analysis, and the authors take full responsibility for all contents.

\subsection{Implementation details of baselines} \label{appendix:baseline-details}
Pretrained checkpoints are used where available; StableVITON is retrained on DressCode upper-body items for consistency. All models use DDIM~\citet{ddim} with $50$ steps and classifier-free guidance (CFG)~\citet{cfg} with scale $1.0$ (except LaDI-VTON, scale $7.5$). For inverse solvers, all methods are adapted to the latent diffusion framework, sharing the same (A) initialization and (C) standard denoising steps ($N=2$), differing only in the (B) measurement-guided sampling component.

\subsection{Inverse solver formulation} \label{appendix:inverse-solver}
\begin{figure*}[t!]
\centering
\includegraphics[width=\textwidth]{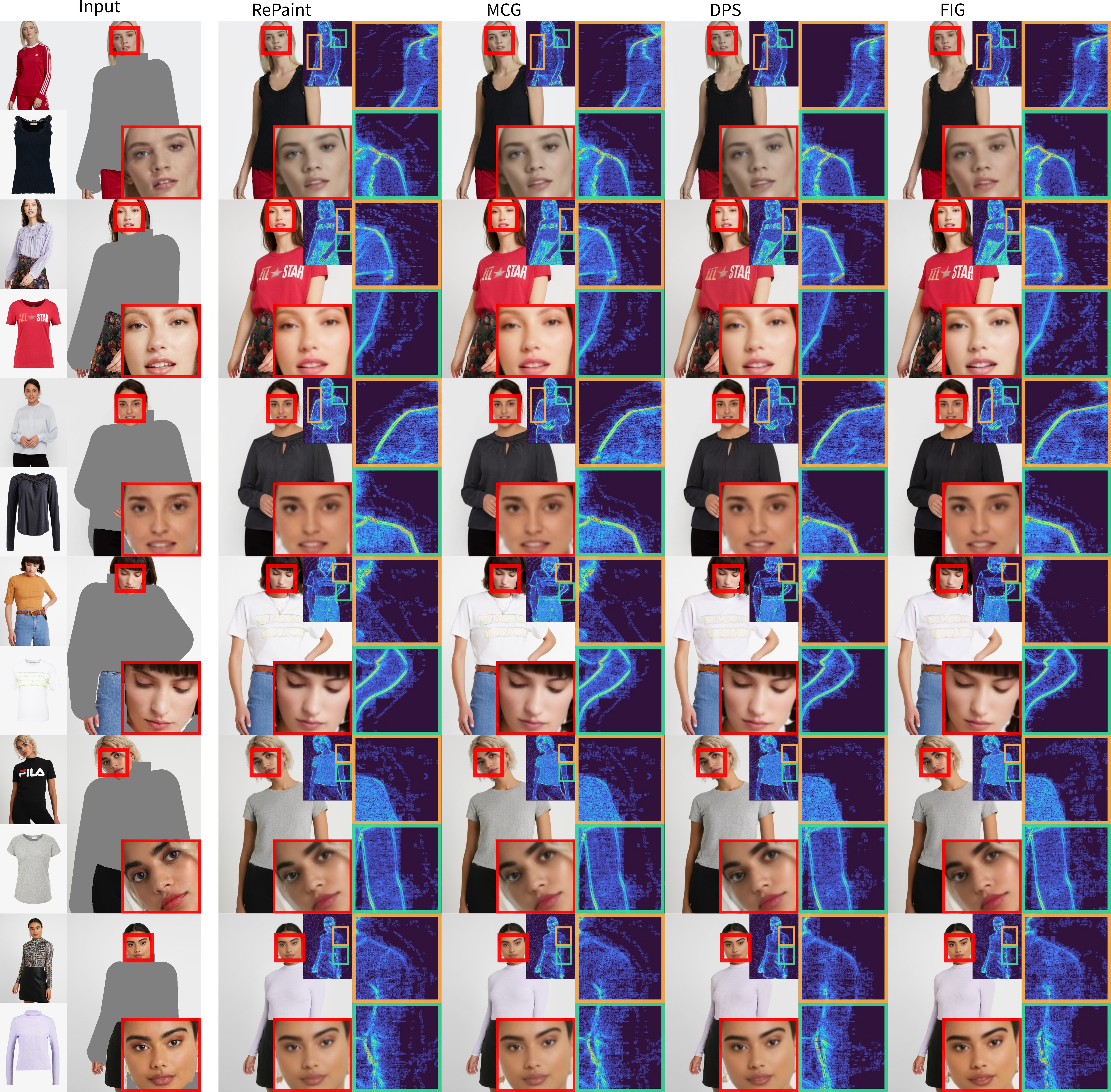} 
\caption{StableVITON on VITON-HD with inverse solvers applied without post-hoc replacement. \textcolor{red}{Red} indicates face zoom-in, and \textcolor{orange}{orange} and \textcolor{green}{green} indicate artifact map zoom-ins. Hard constraint solvers (RePaint, MCG) and progressive update solvers (DPS, FIG) fail to fully satisfy measurements, highlighting the need for post-hoc replacement. Hard constraints generate artifacts due to semantic inconsistencies across regions, whereas progressive updates produce minimal artifacts, as each update induces only small changes.
}
\label{fig:hard}
\end{figure*}

We classify inverse solvers into three types: hard constraints (RePaint~\citet{repaint}, MCG~\citet{mcg}), progressive updates (DPS~\citet{dps}, FIG~\citet{fig}), and hybrid stochastic methods (DreamSampler~\citet{dreamsampler}, TReg~\citet{treg}). Hard constraints induce semantic drift between regions due to strong measurement enforcement, directly causing boundary artifacts. Progressive updates maintain stable optimization and produce minimal artifacts. However, both hard constraints and progressive updates operate in latent space, failing to fully satisfy measurements (Fig.~\ref{fig:hard}). To address this, we apply post-hoc replacement, which can still cause boundary artifacts due to semantic mismatch and spatial discontinuities. Hybrid stochastic methods enforce measurement constraints in pixel space and inject stochastic noise to harmonize regions, reducing artifacts. Nevertheless, persistent semantic drift still leads to artifact formation.

\begin{figure*}[t!]
\centering
\includegraphics[width=\textwidth]{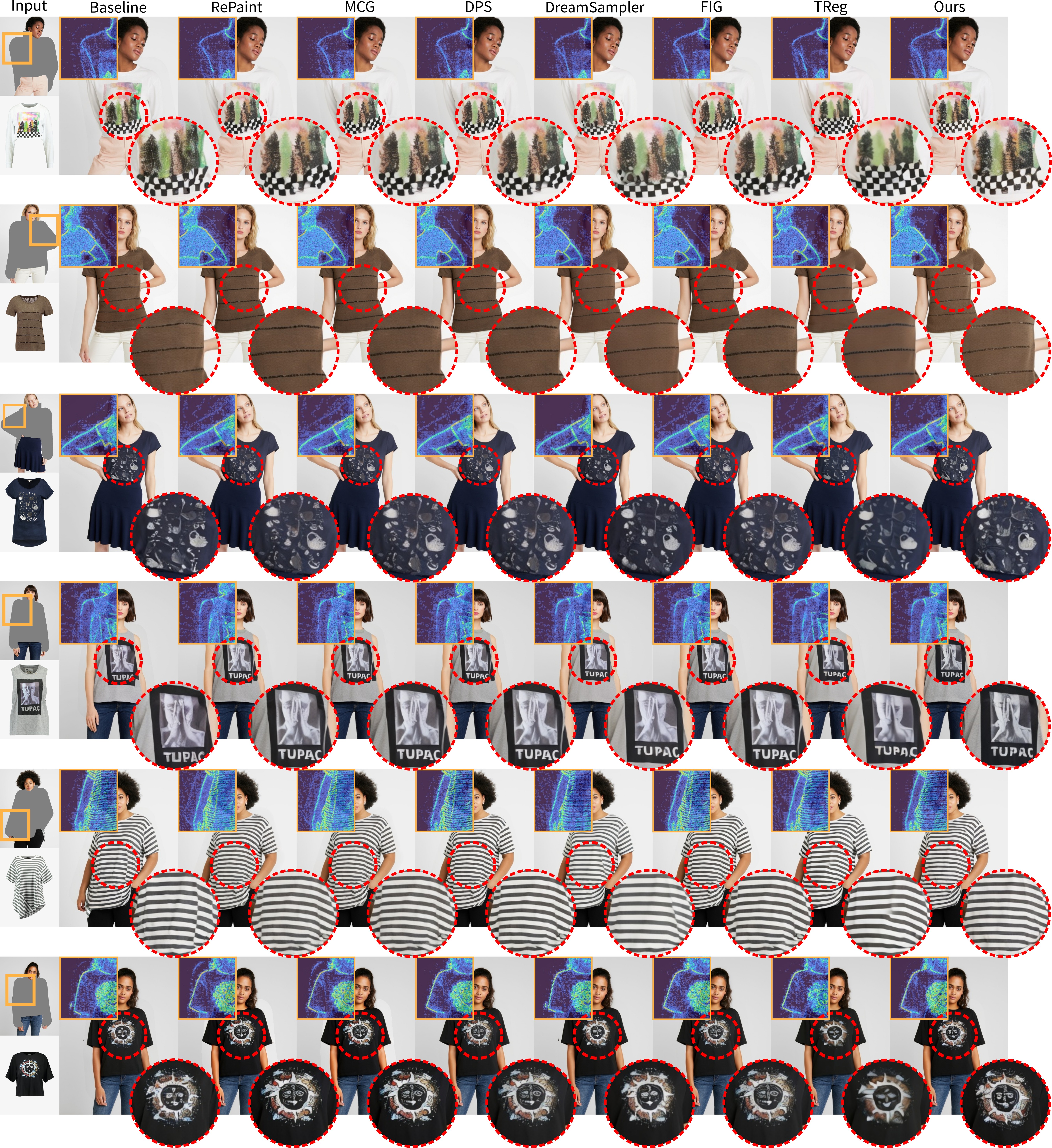} 
\caption{Comparison on the VITON-HD dataset with baseline (StableVITON) and existing inverse solvers. \textcolor{red}{Red} circles highlight texture degradation, particularly in hybrid stochastic methods (DreamSampler, TReg), while our approach preserves fine garment details and patterns. \textcolor{orange}{Orange} boxes indicate artifacts present in other methods, which are absent in our results.
}
\label{fig:ab-inverse-solver}
\end{figure*}

We formulate virtual try-on as an inverse problem and integrate various solver strategies into the latent diffusion sampling process. This section presents the mathematical foundations and implementation details of each approach. We denote the measurement mask as $\mathbf{M}$ and the target measurement as $\mathbf{y}$. The bar notation indicates resizing to match the latent code resolution. Specifically, $\bar{\mathbf{M}}$ denotes the measurement mask with value $1$ in the resized measurement region, and $\bar{\mathbf{y}}$ represents the resized target measurement.
A comparison with the inverse solvers is shown in Fig.~\ref{fig:ab-inverse-solver}.

\noindent \textbf{DDIM sampling~\citet{ddim}.}
The deterministic DDIM sampling forms the basis for all inverse solvers. Given a noisy latent $\mathbf{z}_t$ at timestep $t$, we first estimate the clean latent using Tweedie's formula:
\begin{align}
    \hat{\mathbf{z}}_0^{(t)} = \frac{1}{\sqrt{\bar{\alpha}_t}} \left( \mathbf{z}_t - \sqrt{1 - \bar{\alpha}_t} \cdot \boldsymbol{\epsilon}_\theta(\mathbf{z}_t, t, \mathbf{c}) \right).
\end{align}

The denoising step then updates the latent to timestep $t-1$:
\begin{align}
    \mathbf{z}_{t-1} = \sqrt{\bar{\alpha}_{t-1}}\hat{\mathbf{z}}_0^{(t)} + \sqrt{1-\bar{\alpha}_{t-1}}\boldsymbol{\epsilon}_{\theta}(\mathbf{z}_t,t,\mathbf{c}).
\end{align}

\subsubsection{Hard measurement methods}
These methods enforce measurement consistency through direct projection or replacement in the latent space. 

\noindent \textbf{RePaint~\citet{repaint}.}
This approach replaces the measurement region with noisy observations at each denoising step. We omit the resampling strategy proposed in Repaint as it is too time-consuming:
\begin{align}
    \bar{\mathbf{y}}_{t-1} & \sim \mathcal{N}(\sqrt{\bar{\alpha}_{t-1}}\bar{\mathbf{y}},(1-\bar{\alpha}_{t-1})\mathbf{I}), \\
    \mathbf{z}_{t-1}' &= \bar{\mathbf{M}}\odot\bar{\mathbf{y}}_{t-1} + (1-\bar{\mathbf{M}})\odot\mathbf{z}_{t-1}.
\end{align}

\noindent \textbf{MCG (Manifold-Constrained Gradient)~\citet{mcg}.} This method combines gradient-based optimization with hard projection:
\begin{align}
    \mathbf{z}_{t-1}' &= \mathbf{z}_{t-1} - \gamma \nabla_{\mathbf{z}_{t}} \|\bar{\mathbf{y}}-\bar{\mathbf{M}}\odot\hat{\mathbf{z}}_0^{(t)}\|_2^2, \\
    \mathbf{z}_{t-1}'' &= \bar{\mathbf{M}}\odot \bar{\mathbf{y}}_{t-1} + (1-\bar{\mathbf{M}})\odot\mathbf{z}_{t-1}',
\end{align}
where $\gamma$ is the gradient step size, which we set to $1$.
\subsubsection{Progressive update methods}
These methods guide the sampling trajectory iteratively through gradient updates without relying on hard measurement constraints.

\noindent \textbf{DPS (Diffusion Posterior Sampling)~\citet{dps}.} DPS adjusts the sampling trajectory via measurement consistency gradients computed in the Tweedie space:
\begin{align}
    \mathbf{z}_{t-1}' = \mathbf{z}_{t-1} - \gamma \nabla_{\mathbf{z}_{t}} \|\bar{\mathbf{y}}-\bar{\mathbf{M}}\odot\hat{\mathbf{z}}_0^{(t)}\|_2^2,
\end{align}
where we set $\gamma=1$.

\noindent \textbf{FIG (Flow with Interpolant Guidance)~\citet{fig}.}
By operating directly on the noisy latent, FIG performs gradient updates along the diffusion trajectory, preserving stability and sample diversity, whereas Tweedie-space optimization is more precise but incurs higher computational cost and reduces diversity.
\begin{align}
    \mathbf{z}_{t-1}' = \mathbf{z}_{t-1} - \gamma \nabla_{\mathbf{z}_{t-1}} \|\bar{\mathbf{y}}_{t-1}-\bar{\mathbf{M}}\odot\mathbf{z}_{t-1}\|_2^2,
\end{align}
with $\gamma=1$.

\subsubsection{Hybrid stochastic methods}
These approaches combine deterministic updates with stochastic noise injection, where the degree of stochasticity is controlled through $\eta \beta_t$, to balance measurement consistency and generation diversity.
\begin{align}
    \tilde{\boldsymbol{\epsilon}}_t := \frac{\sqrt{1 - \bar{\alpha}_{t-1} - \eta^2 \beta_t^2} \cdot \boldsymbol{\epsilon}_{\theta} + \eta \beta_t \cdot \boldsymbol{\epsilon}}{\sqrt{1 - \bar{\alpha}_{t-1}}}, \quad \boldsymbol{\epsilon} \sim \mathcal{N}(0,\mathbf{I}),
\end{align}
where $\eta$ controls the noise level and $\beta_t$ is the noise schedule. The pixel-space optimization is performed via gradient updates with a learning rate of $1e{-}3$, a regularization coefficient $\lambda$ of $1e{-}4$, and $1000$ iterations:

\noindent \textbf{DreamSampler~\citet{dreamsampler}.} 
DreamSampler integrates pixel-space and latent-space optimization to guide the diffusion sampling trajectory while maintaining measurement consistency. Let $\varnothing$ denote a null embedding, as introduced in the classifier-free guidance (CFG) framework, used to perform latent optimization without conditioning information. In the final latent update, the stochastic noise term $\tilde{\boldsymbol{\epsilon}}_t$ is set by $\eta\beta_t = \sqrt{\bar{\alpha}_{t}(1-\bar{\alpha}_{t-1})}$, controlling the amount of injected noise to balance diversity and trajectory stability.
\begin{align}
    &\hat{\mathbf{z}}_{0,\varnothing}^{(t)}  =   \frac{1}{\sqrt{\bar{\alpha}_t}} \left( \mathbf{z}_t - \sqrt{1 - \bar{\alpha}_t} \cdot \boldsymbol{\epsilon}_\theta(\mathbf{z}_t, t, \varnothing) \right), \\
    \hat{\mathbf{x}}_{\mathbf{y},\varnothing} =   \argmin_{\mathbf{x}_\varnothing} & \left(     
    \|\mathbf{y} - \bar{\mathbf{M}}\odot\mathbf{x}_\varnothing\|_2^2
    + \lambda \|\mathbf{x}_\varnothing - \mathcal{D}(\hat{\mathbf{z}}_{0,\varnothing}^{(t)})\|_2^2
    \right), \quad \hat{\mathbf{z}}_{\mathbf{y},\varnothing} = \mathcal{E}(\hat{\mathbf{x}}_{\mathbf{y},\varnothing}),  \\
    &\hat{\mathbf{z}}_0^{(t)}(\bar{\alpha}_{t-1}) = \bar{\alpha}_{t-1}\hat{\mathbf{z}}_{\mathbf{y},\varnothing} + (1-\bar{\alpha}_{t-1})\hat{\mathbf{z}}_{0,\varnothing}^{(t)}, \\
    \hat{\mathbf{z}}_0^{(t)}(\bar{\alpha}_{t},\bar{\alpha}_{t-1}) =& \bar{\mathbf{M}}\odot \hat{\mathbf{z}}_0^{(t)}(\bar{\alpha}_{t-1}) + (1-\bar{\mathbf{M}})\odot(\bar{\alpha}_t\hat{\mathbf{z}}_0^{(t)}+(1-\bar{\alpha}_t)\hat{\mathbf{z}}_0^{(t)}(\bar{\alpha}_{t-1})),
    \\
    &\mathbf{z}_{t-1}' = \sqrt{\bar{\alpha}_{t-1}}\hat{\mathbf{z}}_0^{(t)}(\bar{\alpha}_{t}, \bar{\alpha}_{t-1}) + \sqrt{1-\bar{\alpha}_{t-1}}\tilde{\boldsymbol{\epsilon}}_t,
\end{align}
where $\mathcal{E}$ and $\mathcal{D}$ denote encoder and decoder, and $\lambda$ balances data fidelity.

\noindent \textbf{TReg~\citet{treg}.} 
TReg performs the hybrid approach by performing optimization directly in pixel space with latent regularization. It solves a regularized inverse problem where the measurement operator $\bar{\mathbf{M}}$ enforces constraints, while the regularization term maintains semantic coherence via the diffusion prior. In the stochastic update, the noise parameter is set as $\eta\beta_t = \sqrt{\bar{\alpha}_{t-1}(1-\bar{\alpha}_{t-1})}$, following a noise schedule distinct from DreamSampler.
\begin{align}
    \hat{\mathbf{x}}_{\mathbf{y}} = \argmin_{\mathbf{x}} & \left(    
    \|\mathbf{y} - \bar{\mathbf{M}}\odot\mathbf{x}\|_2^2
    + \lambda \|\mathbf{x} - \mathcal{D}(\hat{\mathbf{z}}_0^{(t)})\|_2^2
    \right), \quad \hat{\mathbf{z}}_{\mathbf{y}} = \mathcal{E}(\hat{\mathbf{x}}_{\mathbf{y}}),  \\
    \hat{\mathbf{z}}_0^{(t)}&(\bar{\alpha}_{t-1}) = \bar{\alpha}_{t-1} \hat{\mathbf{z}}_{\mathbf{y}} + (1- \bar{\alpha}_{t-1})\hat{\mathbf{z}}_0^{(t)}, \\
    \mathbf{z}_{t-1}' &= \sqrt{\bar{\alpha}_{t-1}}\hat{\mathbf{z}}_0^{(t)}(\bar{\alpha}_{t-1}) + \sqrt{1-\bar{\alpha}_{t-1}}\tilde{\boldsymbol{\epsilon}}_t.
\end{align}
\subsection{Additional results}

\begin{figure*}[t!]
\centering
\includegraphics[width=\textwidth]{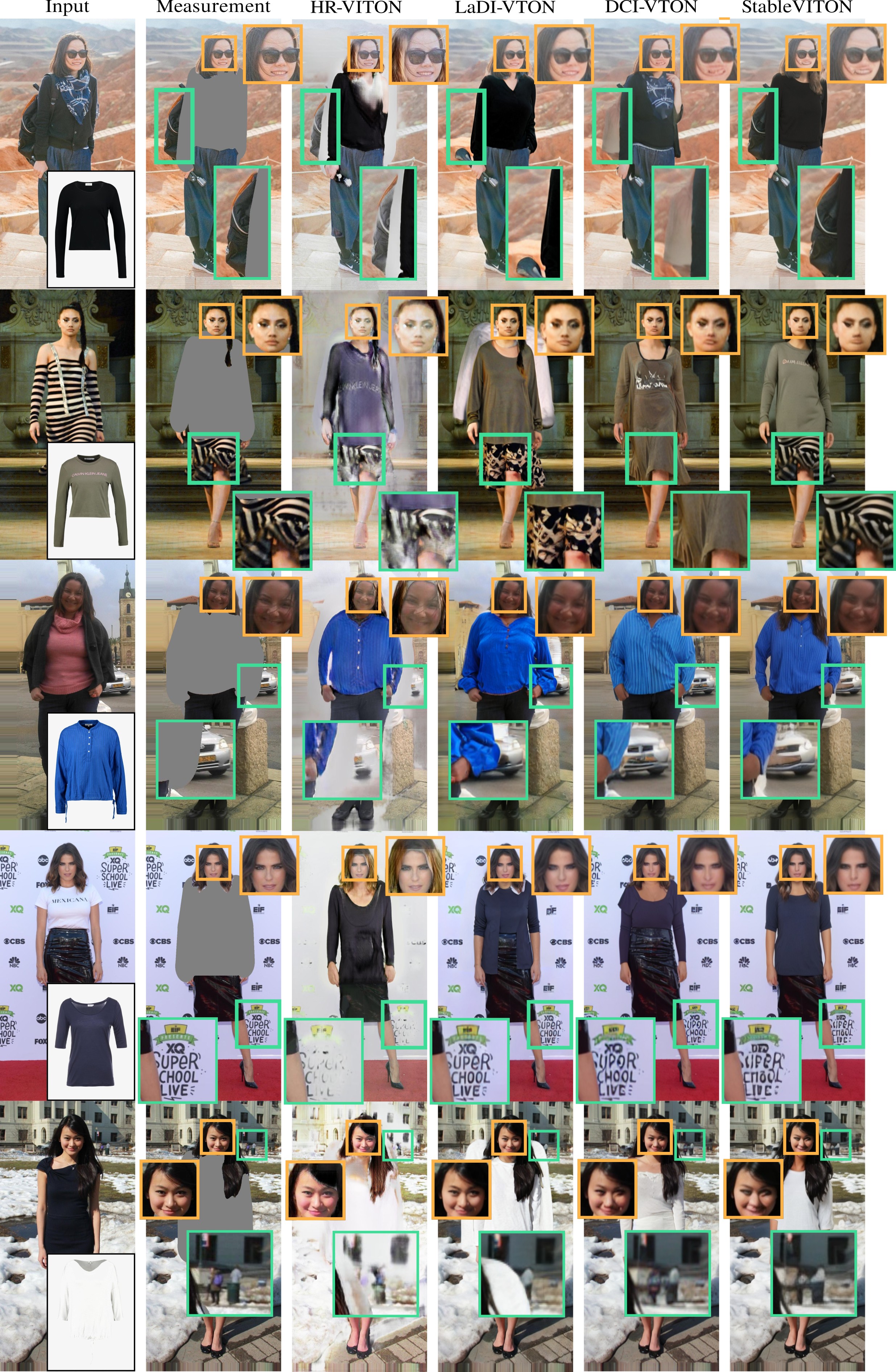} 
\caption{Qualitative results of baseline models on the SHHQ-1.0 dataset. Our observations show that generated images fail to preserve content in non-try-on regions: bags, skirts, cars, text, and human features (\textcolor{green}{green} boxes). \textcolor{orange}{Orange} boxes indicate areas where facial details are not properly preserved.
}
\label{fig:ab-non-try-on}
\end{figure*}

\begin{figure*}[t!]
\vspace{-0.1cm}
\centering
\includegraphics[width=\linewidth]{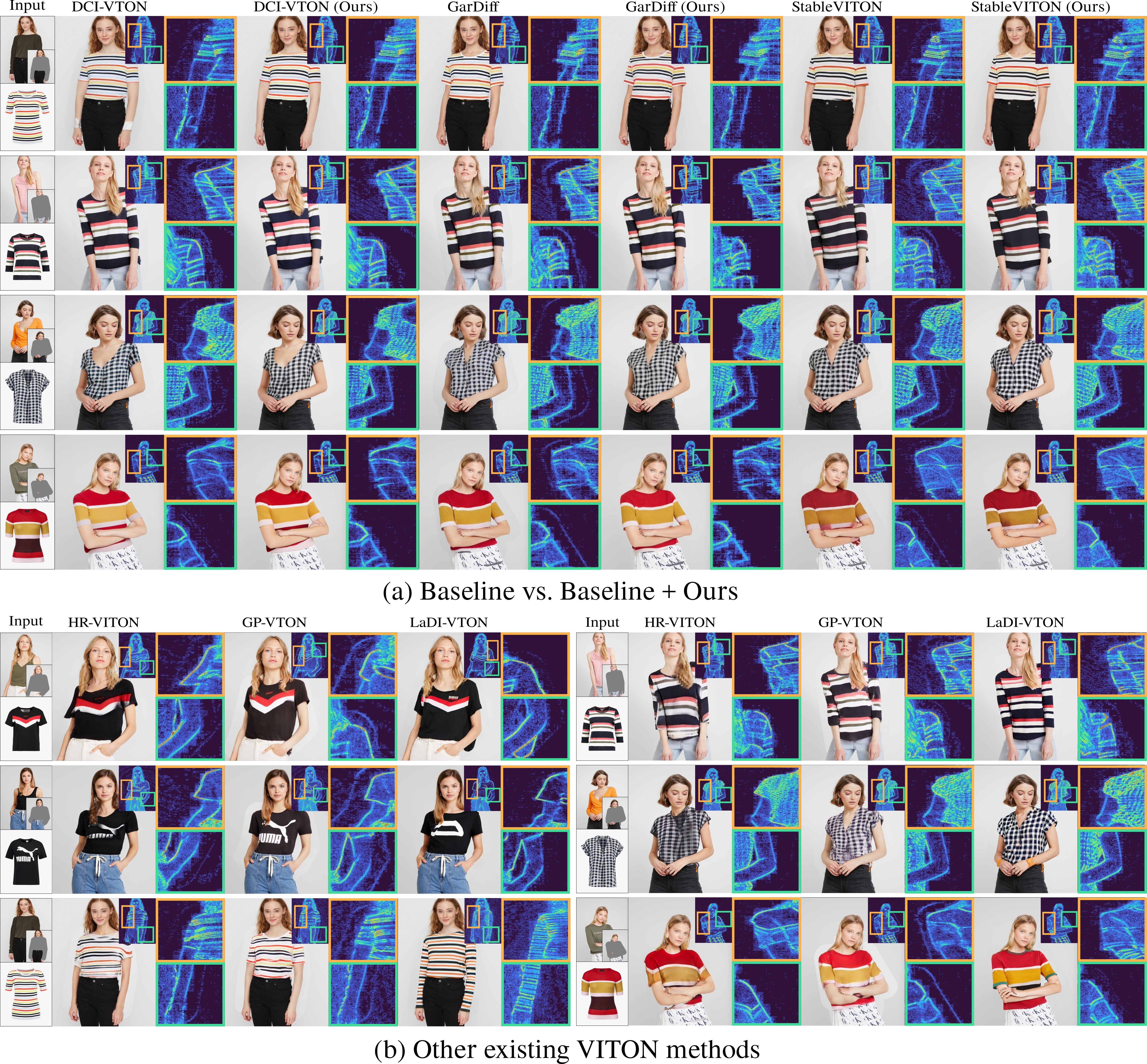} 
\vspace{-0.05\textwidth}
\caption{Additional qualitative results on the VITON-HD comparing baseline methods with our approach. (a) Comparison of baselines and their versions enhanced with our method: our approach consistently removes boundary artifacts while preserving high-frequency garment details such as logos, text, and complex patterns. (b) Results of the remaining models without our enhancement: in 2-stage pipeline models, warping results show garment distortions and color inconsistencies.
}
\label{fig:ab-viton}
\end{figure*}


\begin{figure*}[t!]
\centering
\includegraphics[width=\linewidth]{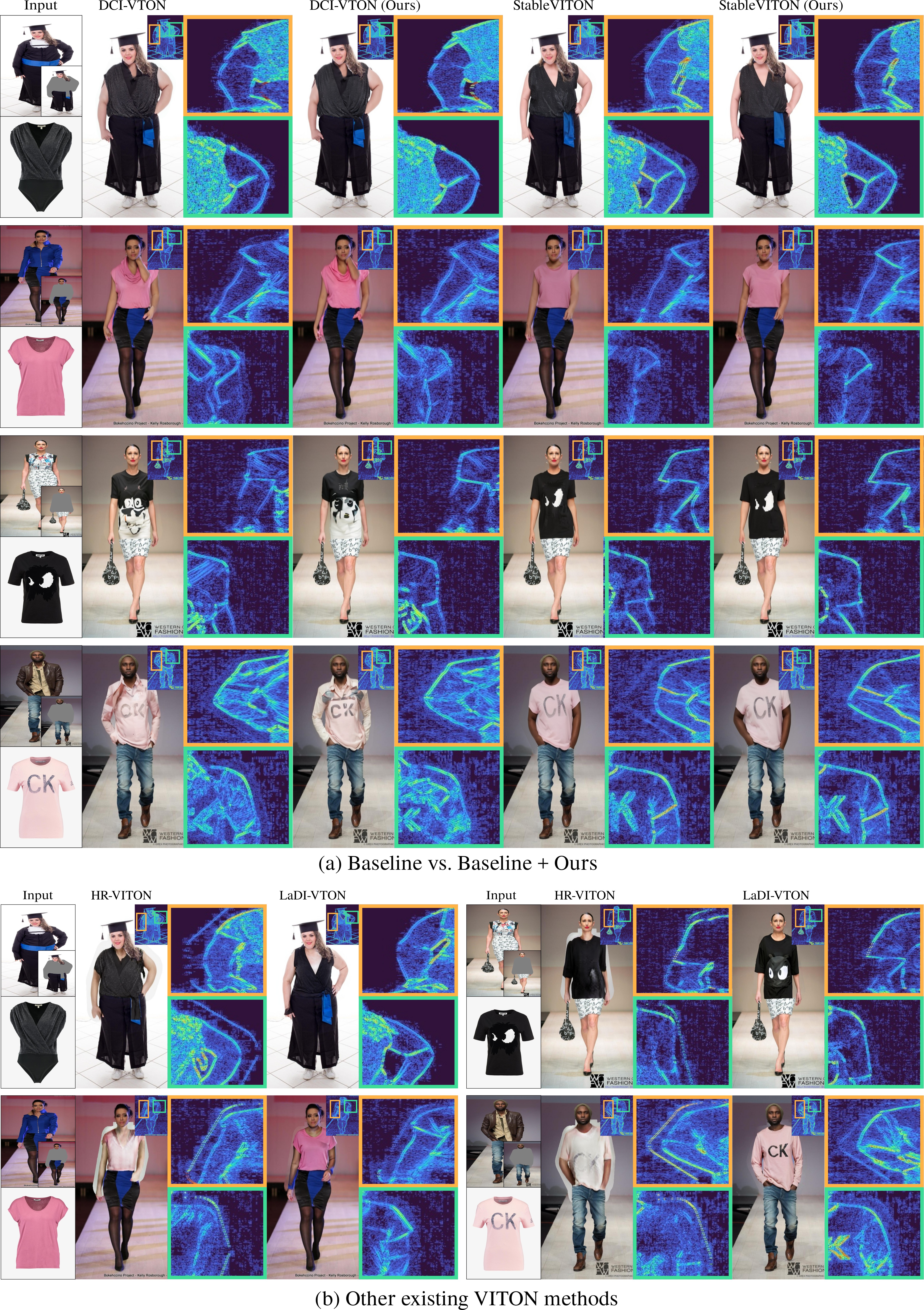} 
\caption{Extended comparison demonstrating robustness across domains on the SHHQ-1.0 dataset. (a) Comparison of baselines and their versions enhanced with our method: even in cross-domain scenarios, our approach effectively removes artifacts, demonstrating robustness. (b) Other VITON methods show boundary artifacts and garment distortion, whereas our approach preserves boundaries and garment details.
}
\label{fig:ab-shhq}
\end{figure*}





\begin{figure*}[t!]
\centering
\includegraphics[width=\textwidth]{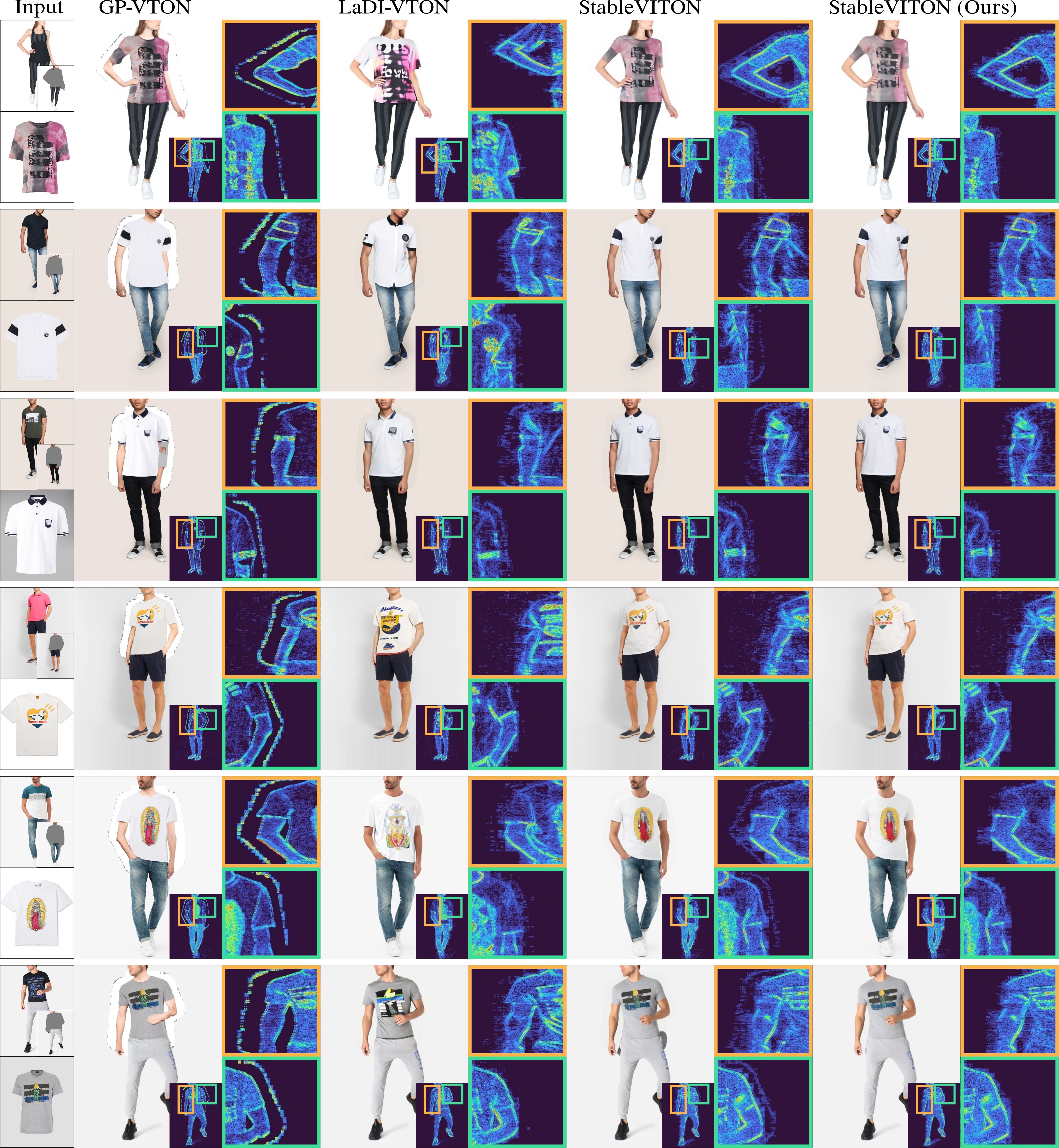} 
\caption{Qualitative comparison of baseline VITON methods on DressCode dataset. Traditional methods (GP-VTON, LaDI-VTON) exhibit misalignment and texture distortion, while StableVITON shows boundary artifacts despite better garment alignment. Our method applied to StableVITON (rightmost) eliminates boundary inconsistencies while preserving both garment details and identity features.
}
\label{fig:ab-dress}
\end{figure*}

%% file: main.bbl
\begin{thebibliography}{48}
\providecommand{\natexlab}[1]{#1}
\providecommand{\url}[1]{\texttt{#1}}
\expandafter\ifx\csname urlstyle\endcsname\relax
  \providecommand{\doi}[1]{doi: #1}\else
  \providecommand{\doi}{doi: \begingroup \urlstyle{rm}\Url}\fi

\bibitem[Almog et~al.(2025)Almog, Shamir, and Fried]{reed-vae}
Gal Almog, Ariel Shamir, and Ohad Fried.
\newblock Reed-vae: Re-encode decode training for iterative image editing with diffusion models.
\newblock In \emph{Computer Graphics Forum}, pp.\  e70020. Wiley Online Library, 2025.

\bibitem[Choi et~al.(2021{\natexlab{a}})Choi, Kim, Jeong, Gwon, and Yoon]{ilvr}
Jooyoung Choi, Sungwon Kim, Yonghyun Jeong, Youngjune Gwon, and Sungroh Yoon.
\newblock Ilvr: Conditioning method for denoising diffusion probabilistic models.
\newblock In \emph{Proceedings of the IEEE/CVF International Conference on Computer Vision}, pp.\  14367--14376, 2021{\natexlab{a}}.

\bibitem[Choi et~al.(2022)Choi, Lee, Shin, Kim, Kim, and Yoon]{perception}
Jooyoung Choi, Jungbeom Lee, Chaehun Shin, Sungwon Kim, Hyunwoo Kim, and Sungroh Yoon.
\newblock Perception prioritized training of diffusion models.
\newblock In \emph{Proceedings of the IEEE/CVF conference on Computer Vision and Pattern Recognition}, pp.\  11472--11481, 2022.

\bibitem[Choi et~al.(2021{\natexlab{b}})Choi, Park, Lee, and Choo]{vitonhd}
Seunghwan Choi, Sunghyun Park, Minsoo Lee, and Jaegul Choo.
\newblock Viton-hd: High-resolution virtual try-on via misalignment-aware normalization.
\newblock In \emph{Proceedings of the IEEE/CVF conference on computer vision and pattern recognition}, pp.\  14131--14140, 2021{\natexlab{b}}.

\bibitem[Choi et~al.(2024)Choi, Kwak, Lee, Choi, and Shin]{idmvton}
Yisol Choi, Sangkyung Kwak, Kyungmin Lee, Hyungwon Choi, and Jinwoo Shin.
\newblock Improving diffusion models for authentic virtual try-on in the wild.
\newblock In \emph{European Conference on Computer Vision}, pp.\  206--235. Springer, 2024.

\bibitem[Chung et~al.(2022)Chung, Sim, Ryu, and Ye]{mcg}
Hyungjin Chung, Byeongsu Sim, Dohoon Ryu, and Jong~Chul Ye.
\newblock Improving diffusion models for inverse problems using manifold constraints.
\newblock \emph{Advances in Neural Information Processing Systems}, 35:\penalty0 25683--25696, 2022.

\bibitem[Chung et~al.(2023)Chung, Kim, Mccann, Klasky, and Ye]{dps}
Hyungjin Chung, Jeongsol Kim, Michael~Thompson Mccann, Marc~Louis Klasky, and Jong~Chul Ye.
\newblock Diffusion posterior sampling for general noisy inverse problems.
\newblock In \emph{The Eleventh International Conference on Learning Representations}, 2023.

\bibitem[Chung et~al.(2024)Chung, Ye, Milanfar, and Delbracio]{promptsolver}
Hyungjin Chung, Jong~Chul Ye, Peyman Milanfar, and Mauricio Delbracio.
\newblock Prompt-tuning latent diffusion models for inverse problems.
\newblock In \emph{Proceedings of the 41st International Conference on Machine Learning}, pp.\  8941--8967, 2024.

\bibitem[Fu et~al.(2022)Fu, Li, Jiang, Lin, Qian, Loy, Wu, and Liu]{shhq}
Jianglin Fu, Shikai Li, Yuming Jiang, Kwan-Yee Lin, Chen Qian, Chen-Change Loy, Wayne Wu, and Ziwei Liu.
\newblock Stylegan-human: A data-centric odyssey of human generation.
\newblock \emph{arXiv preprint}, arXiv:2204.11823, 2022.

\bibitem[Gal et~al.(2023)Gal, Alaluf, Atzmon, Patashnik, Bermano, Chechik, and Cohen-or]{textual-inversion}
Rinon Gal, Yuval Alaluf, Yuval Atzmon, Or~Patashnik, Amit~Haim Bermano, Gal Chechik, and Daniel Cohen-or.
\newblock An image is worth one word: Personalizing text-to-image generation using textual inversion.
\newblock In \emph{The Eleventh International Conference on Learning Representations}, 2023.

\bibitem[Ge et~al.(2021)Ge, Song, Zhang, Ge, Liu, and Luo]{parser}
Yuying Ge, Yibing Song, Ruimao Zhang, Chongjian Ge, Wei Liu, and Ping Luo.
\newblock Parser-free virtual try-on via distilling appearance flows.
\newblock In \emph{Proceedings of the IEEE/CVF conference on computer vision and pattern recognition}, pp.\  8485--8493, 2021.

\bibitem[Gou et~al.(2023)Gou, Sun, Zhang, Si, Qian, and Zhang]{dcivton}
Junhong Gou, Siyu Sun, Jianfu Zhang, Jianlou Si, Chen Qian, and Liqing Zhang.
\newblock Taming the power of diffusion models for high-quality virtual try-on with appearance flow.
\newblock In \emph{Proceedings of the 31st ACM International Conference on Multimedia}, pp.\  7599--7607, 2023.

\bibitem[Han et~al.(2018)Han, Wu, Wu, Yu, and Davis]{viton}
Xintong Han, Zuxuan Wu, Zhe Wu, Ruichi Yu, and Larry~S Davis.
\newblock Viton: An image-based virtual try-on network.
\newblock In \emph{Proceedings of the IEEE conference on computer vision and pattern recognition}, pp.\  7543--7552, 2018.

\bibitem[Ho \& Salimans(2022)Ho and Salimans]{cfg}
Jonathan Ho and Tim Salimans.
\newblock Classifier-free diffusion guidance.
\newblock \emph{arXiv preprint arXiv:2207.12598}, 2022.

\bibitem[Ho et~al.(2020)Ho, Jain, and Abbeel]{ddpm}
Jonathan Ho, Ajay Jain, and Pieter Abbeel.
\newblock Denoising diffusion probabilistic models.
\newblock \emph{Advances in neural information processing systems}, 33:\penalty0 6840--6851, 2020.

\bibitem[Hu et~al.(2022)Hu, Shen, Wallis, Allen-Zhu, Li, Wang, Wang, Chen, et~al.]{lora}
Edward~J Hu, Yelong Shen, Phillip Wallis, Zeyuan Allen-Zhu, Yuanzhi Li, Shean Wang, Lu~Wang, Weizhu Chen, et~al.
\newblock Lora: Low-rank adaptation of large language models.
\newblock \emph{ICLR}, 1\penalty0 (2):\penalty0 3, 2022.

\bibitem[Hu(2024)]{animate}
Li~Hu.
\newblock Animate anyone: Consistent and controllable image-to-video synthesis for character animation.
\newblock In \emph{Proceedings of the IEEE/CVF Conference on Computer Vision and Pattern Recognition}, pp.\  8153--8163, 2024.

\bibitem[Kim et~al.(2024{\natexlab{a}})Kim, Gu, Park, Park, and Choo]{stableviton}
Jeongho Kim, Guojung Gu, Minho Park, Sunghyun Park, and Jaegul Choo.
\newblock Stableviton: Learning semantic correspondence with latent diffusion model for virtual try-on.
\newblock In \emph{Proceedings of the IEEE/CVF conference on computer vision and pattern recognition}, pp.\  8176--8185, 2024{\natexlab{a}}.

\bibitem[Kim et~al.(2024{\natexlab{b}})Kim, Park, and Ye]{dreamsampler}
Jeongsol Kim, Geon~Yeong Park, and Jong~Chul Ye.
\newblock Dreamsampler: Unifying diffusion sampling and score distillation for image manipulation.
\newblock In \emph{European Conference on Computer Vision}, pp.\  398--414. Springer, 2024{\natexlab{b}}.

\bibitem[Kim et~al.(2025)Kim, Park, Chung, and Ye]{treg}
Jeongsol Kim, Geon~Yeong Park, Hyungjin Chung, and Jong~Chul Ye.
\newblock Regularization by texts for latent diffusion inverse solvers.
\newblock In \emph{The Thirteenth International Conference on Learning Representations}, 2025.

\bibitem[Kingma \& Welling(2022)Kingma and Welling]{vae}
Diederik~P Kingma and Max Welling.
\newblock Auto-encoding variational bayes, 2022.
\newblock URL \url{https://arxiv.org/abs/1312.6114}.

\bibitem[Lee et~al.(2022)Lee, Gu, Park, Choi, and Choo]{hrviton}
Sangyun Lee, Gyojung Gu, Sunghyun Park, Seunghwan Choi, and Jaegul Choo.
\newblock High-resolution virtual try-on with misalignment and occlusion-handled conditions.
\newblock In \emph{European Conference on Computer Vision}, pp.\  204--219. Springer, 2022.

\bibitem[Lin et~al.(2024)Lin, Liu, Li, and Yang]{common}
Shanchuan Lin, Bingchen Liu, Jiashi Li, and Xiao Yang.
\newblock Common diffusion noise schedules and sample steps are flawed.
\newblock In \emph{Proceedings of the IEEE/CVF winter conference on applications of computer vision}, pp.\  5404--5411, 2024.

\bibitem[Lugmayr et~al.(2022{\natexlab{a}})Lugmayr, Danelljan, Romero, Yu, Timofte, and Gool]{ddim}
Andreas Lugmayr, Martin Danelljan, Andres Romero, Fisher Yu, Radu Timofte, and Luc~Van Gool.
\newblock Repaint: Inpainting using denoising diffusion probabilistic models, 2022{\natexlab{a}}.
\newblock URL \url{https://arxiv.org/abs/2201.09865}.

\bibitem[Lugmayr et~al.(2022{\natexlab{b}})Lugmayr, Danelljan, Romero, Yu, Timofte, and Van~Gool]{repaint}
Andreas Lugmayr, Martin Danelljan, Andres Romero, Fisher Yu, Radu Timofte, and Luc Van~Gool.
\newblock Repaint: Inpainting using denoising diffusion probabilistic models.
\newblock In \emph{Proceedings of the IEEE/CVF conference on computer vision and pattern recognition}, pp.\  11461--11471, 2022{\natexlab{b}}.

\bibitem[Morelli et~al.(2022)Morelli, Fincato, Cornia, Landi, Cesari, and Cucchiara]{dresscode}
Davide Morelli, Matteo Fincato, Marcella Cornia, Federico Landi, Fabio Cesari, and Rita Cucchiara.
\newblock Dress code: High-resolution multi-category virtual try-on.
\newblock In \emph{Proceedings of the IEEE/CVF conference on computer vision and pattern recognition}, pp.\  2231--2235, 2022.

\bibitem[Morelli et~al.(2023)Morelli, Baldrati, Cartella, Cornia, Bertini, and Cucchiara]{ladivton}
Davide Morelli, Alberto Baldrati, Giuseppe Cartella, Marcella Cornia, Marco Bertini, and Rita Cucchiara.
\newblock Ladi-vton: Latent diffusion textual-inversion enhanced virtual try-on.
\newblock In \emph{Proceedings of the 31st ACM international conference on multimedia}, pp.\  8580--8589, 2023.

\bibitem[Novitskiy et~al.(2025)Novitskiy, Vasilev, Kovaleva, Arkhipkin, and Dimitrov]{vivat}
Lev Novitskiy, Viacheslav Vasilev, Maria Kovaleva, Vladimir Arkhipkin, and Denis Dimitrov.
\newblock Vivat: Virtuous improving vae training through artifact mitigation.
\newblock \emph{arXiv preprint arXiv:2506.07863}, 2025.

\bibitem[OpenAI(2025)]{openai2025chatgpt}
OpenAI.
\newblock Chatgpt (gpt-5).
\newblock \url{https://chat.openai.com/}, 2025.
\newblock Large language model.

\bibitem[Podell et~al.(2024)Podell, English, Lacey, Blattmann, Dockhorn, M{\"u}ller, Penna, and Rombach]{sdxl}
Dustin Podell, Zion English, Kyle Lacey, Andreas Blattmann, Tim Dockhorn, Jonas M{\"u}ller, Joe Penna, and Robin Rombach.
\newblock Sdxl: Improving latent diffusion models for high-resolution image synthesis.
\newblock In \emph{The Twelfth International Conference on Learning Representations}, 2024.

\bibitem[Ramesh et~al.(2021)Ramesh, Pavlov, Goh, Gray, Voss, Radford, Chen, and Sutskever]{dalle}
Aditya Ramesh, Mikhail Pavlov, Gabriel Goh, Scott Gray, Chelsea Voss, Alec Radford, Mark Chen, and Ilya Sutskever.
\newblock Zero-shot text-to-image generation.
\newblock In \emph{International conference on machine learning}, pp.\  8821--8831. Pmlr, 2021.

\bibitem[Rombach et~al.(2022)Rombach, Blattmann, Lorenz, Esser, and Ommer]{sd}
Robin Rombach, Andreas Blattmann, Dominik Lorenz, Patrick Esser, and Bj{\"o}rn Ommer.
\newblock High-resolution image synthesis with latent diffusion models.
\newblock In \emph{Proceedings of the IEEE/CVF conference on computer vision and pattern recognition}, pp.\  10684--10695, 2022.

\bibitem[Rout et~al.(2023)Rout, Raoof, Daras, Caramanis, Dimakis, and Shakkottai]{psld}
Litu Rout, Negin Raoof, Giannis Daras, Constantine Caramanis, Alex Dimakis, and Sanjay Shakkottai.
\newblock Solving linear inverse problems provably via posterior sampling with latent diffusion models.
\newblock \emph{Advances in Neural Information Processing Systems}, 36:\penalty0 49960--49990, 2023.

\bibitem[Ruiz et~al.(2023)Ruiz, Li, Jampani, Pritch, Rubinstein, and Aberman]{dreambooth}
Nataniel Ruiz, Yuanzhen Li, Varun Jampani, Yael Pritch, Michael Rubinstein, and Kfir Aberman.
\newblock Dreambooth: Fine tuning text-to-image diffusion models for subject-driven generation.
\newblock In \emph{Proceedings of the IEEE/CVF conference on computer vision and pattern recognition}, pp.\  22500--22510, 2023.

\bibitem[Seyfioglu et~al.(2023)Seyfioglu, Bouyarmane, Kumar, Tavanaei, and Tutar]{dreampaint}
Mehmet~Saygin Seyfioglu, Karim Bouyarmane, Suren Kumar, Amir Tavanaei, and Ismail~B Tutar.
\newblock Dreampaint: Few-shot inpainting of e-commerce items for virtual try-on without 3d modeling.
\newblock \emph{arXiv preprint arXiv:2305.01257}, 2023.

\bibitem[Song et~al.(2024)Song, Kwon, Zhang, Hu, Qu, and Shen]{resample}
Bowen Song, Soo~Min Kwon, Zecheng Zhang, Xinyu Hu, Qing Qu, and Liyue Shen.
\newblock Solving inverse problems with latent diffusion models via hard data consistency, 2024.
\newblock URL \url{https://arxiv.org/abs/2307.08123}.

\bibitem[Wan et~al.(2024)Wan, Li, Chen, Pan, Yao, Cao, and Mei]{gardiff}
Siqi Wan, Yehao Li, Jingwen Chen, Yingwei Pan, Ting Yao, Yang Cao, and Tao Mei.
\newblock Improving virtual try-on with garment-focused diffusion models.
\newblock In \emph{European Conference on Computer Vision}, pp.\  184--199. Springer, 2024.

\bibitem[Wang et~al.(2024)Wang, Chen, Chen, Huang, Jiang, Wang, and Shan]{fldmvton}
Chenhui Wang, Tao Chen, Zhihao Chen, Zhizhong Huang, Taoran Jiang, Qi~Wang, and Hongming Shan.
\newblock Fldm-vton: Faithful latent diffusion model for virtual try-on.
\newblock In \emph{IJCAI}, 2024.

\bibitem[Wu et~al.(2024)Wu, Yang, Sun, Zhang, Li, and Zhang]{seesr}
Rongyuan Wu, Tao Yang, Lingchen Sun, Zhengqiang Zhang, Shuai Li, and Lei Zhang.
\newblock Seesr: Towards semantics-aware real-world image super-resolution.
\newblock In \emph{Proceedings of the IEEE/CVF conference on computer vision and pattern recognition}, pp.\  25456--25467, 2024.

\bibitem[Xie et~al.(2023)Xie, Huang, Dong, Zhao, Dong, Zhang, Zhu, and Liang]{gpvton}
Zhenyu Xie, Zaiyu Huang, Xin Dong, Fuwei Zhao, Haoye Dong, Xijin Zhang, Feida Zhu, and Xiaodan Liang.
\newblock Gp-vton: Towards general purpose virtual try-on via collaborative local-flow global-parsing learning.
\newblock In \emph{Proceedings of the IEEE/CVF conference on computer vision and pattern recognition}, pp.\  23550--23559, 2023.

\bibitem[Yan et~al.(2025)Yan, Zhang, Meng, and Zhao]{fig}
Yici Yan, Yichi Zhang, Xiangming Meng, and Zhizhen Zhao.
\newblock Fig: Flow with interpolant guidance for linear inverse problems.
\newblock In \emph{The Thirteenth International Conference on Learning Representations}, 2025.

\bibitem[Yang et~al.(2020)Yang, Zhang, Guo, Liu, Zuo, and Luo]{cpvton}
Han Yang, Ruimao Zhang, Xiaobao Guo, Wei Liu, Wangmeng Zuo, and Ping Luo.
\newblock Towards photo-realistic virtual try-on by adaptively generating-preserving image content.
\newblock In \emph{Proceedings of the IEEE/CVF conference on computer vision and pattern recognition}, pp.\  7850--7859, 2020.

\bibitem[Yang et~al.(2023)Yang, Wu, Ren, Xie, and Zhang]{pasd}
Tao Yang, Rongyuan Wu, Peiran Ren, Xuansong Xie, and Lei Zhang.
\newblock Pixel-aware stable diffusion for realistic image super-resolution and personalized stylization.
\newblock In \emph{The European Conference on Computer Vision (ECCV) 2024}, 2023.

\bibitem[Ye et~al.(2023)Ye, Zhang, Liu, Han, and Yang]{ip-adapter}
Hu~Ye, Jun Zhang, Sibo Liu, Xiao Han, and Wei Yang.
\newblock Ip-adapter: Text compatible image prompt adapter for text-to-image diffusion models.
\newblock \emph{arXiv preprint arXiv:2308.06721}, 2023.

\bibitem[Yu et~al.(2019)Yu, Wang, and Xie]{vtnfp}
Ruiyun Yu, Xiaoqi Wang, and Xiaohui Xie.
\newblock Vtnfp: An image-based virtual try-on network with body and clothing feature preservation.
\newblock In \emph{Proceedings of the IEEE/CVF international conference on computer vision}, pp.\  10511--10520, 2019.

\bibitem[Zhang et~al.(2025{\natexlab{a}})Zhang, Huang, Liu, Guo, and Huang]{diff-4k}
Jinjin Zhang, Qiuyu Huang, Junjie Liu, Xiefan Guo, and Di~Huang.
\newblock Diffusion-4k: Ultra-high-resolution image synthesis with latent diffusion models.
\newblock In \emph{Proceedings of the Computer Vision and Pattern Recognition Conference}, pp.\  23464--23473, 2025{\natexlab{a}}.

\bibitem[Zhang et~al.(2023)Zhang, Rao, and Agrawala]{controlnet}
Lvmin Zhang, Anyi Rao, and Maneesh Agrawala.
\newblock Adding conditional control to text-to-image diffusion models.
\newblock In \emph{Proceedings of the IEEE/CVF international conference on computer vision}, pp.\  3836--3847, 2023.

\bibitem[Zhang et~al.(2025{\natexlab{b}})Zhang, Song, Zhan, Chang, Zeng, Chen, Luo, and Liu]{boowvton}
Xuanpu Zhang, Dan Song, Pengxin Zhan, Tianyu Chang, Jianhao Zeng, Qingguo Chen, Weihua Luo, and An-An Liu.
\newblock Boow-vton: Boosting in-the-wild virtual try-on via mask-free pseudo data training.
\newblock In \emph{Proceedings of the Computer Vision and Pattern Recognition Conference}, pp.\  26399--26408, 2025{\natexlab{b}}.

\end{thebibliography}
